\PassOptionsToPackage{table}{xcolor}
\documentclass[10pt,twocolumn,letterpaper]{article}
\usepackage[T1]{fontenc}

\usepackage{cvpr}

\definecolor{cvprblue}{rgb}{0.21,0.49,0.74}
\usepackage[pagebackref,breaklinks,colorlinks,allcolors=cvprblue]{hyperref}
\usepackage{multirow}
\usepackage{pifont}

\def\confName{CVPR}
\def\confYear{2027}

\title{Beyond Heuristics: Learning to Compress Visual Tokens\\for Video Large Language Models}

\author{
Xiao Zhang$^{1,2}$,
Wang Zeng$^2$,
Sheng Jin$^{2*}$,
Yunzhe Liu$^1$,
Wentao Liu$^2$,
Chen Qian$^2$,
Shichao Kan$^{1\dagger}$\\
$^1$School of Computer Science and Engineering, Central South University\\
$^2$SenseTime Research and Tetras.AI\\
{\tt\small
xiaozhang@csu.edu.cn \quad
jinsheng@tetras.ai \quad
kanshichao@csu.edu.cn
}
}

\begin{document}
\maketitle

\begingroup
\renewcommand{\thefootnote}{}
\footnotetext{\textsuperscript{*}Project leader. \quad
\textsuperscript{$\dagger$}Corresponding author.}
\endgroup

\begin{abstract}
Video large language models (Video LLMs) incur substantial computational overhead from excessive visual tokens. Most existing video token compression methods rely on training-free, hand-crafted heuristics, limiting their adaptability to diverse video content. Although recent learnable approaches introduce trainable mechanisms for token compression, they often remain constrained by predefined compression rules or require additional alignment training.
To address these limitations, we propose \textbf{COMPACT}, a novel learnable framework for video token \textbf{COM}pression via \textbf{P}rototype-guided \textbf{A}daptive \textbf{C}lus\textbf{T}ering. Specifically, COMPACT first constructs representative visual prototypes through position-aware cross-frame clustering and subsequently refines them through attention-based interactions among prototypes and with the dense visual tokens. Based on the refined prototypes, COMPACT combines the initial token-to-prototype similarities with learned affinities to obtain differentiable soft assignments, which guide the weighted aggregation of original visual tokens into compact representations. By retaining the original visual feature space, this design eliminates the need for additional alignment training.
Extensive experiments on five video understanding benchmarks and three representative Video LLMs demonstrate that COMPACT consistently achieves state-of-the-art performance across various visual token retention ratios, substantially outperforming existing methods, particularly under extremely aggressive compression. Our code is available at \url{https://github.com/xiaozhang79/COMPACT}.
\end{abstract}
\section{Introduction}
\label{sec:intro}

Recent video large language models (Video LLMs)~\cite{li2025llava,zhang2025llava,bai2025qwen2,bai2025qwen3,zhu2025internvl3,yan2026internvideo3} have substantially improved video understanding and reasoning capabilities, enabling models to interpret complex temporal dynamics and answer diverse questions about video content. However, processing videos introduces substantially higher computational costs than processing static images. To capture temporal information, Video LLMs typically sample multiple frames and encode them into long sequences of visual tokens, whose number grows with both the number of frames and spatial resolution. These excessive visual tokens substantially increase the computational cost of LLM prefill and the memory consumption of the KV cache, posing significant challenges to efficient video understanding, particularly for long videos. Meanwhile, videos inherently contain substantial spatiotemporal redundancy, as similar visual content often appears across different frames. Therefore, efficiently compressing visual tokens while preserving informative content for downstream video understanding and reasoning is an important research direction.

\begin{figure*}[t]
\centering
\includegraphics[width=\textwidth]{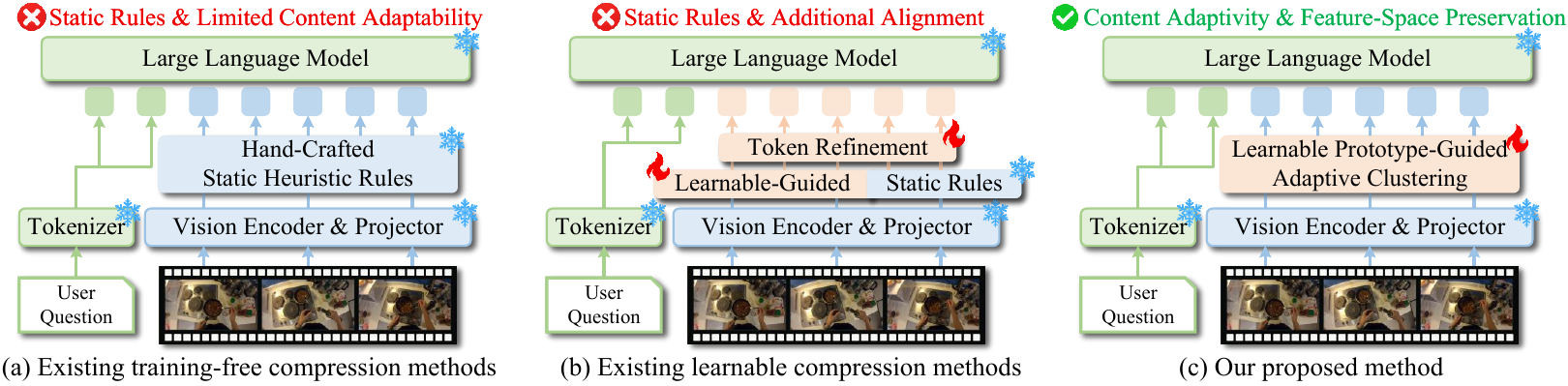}
\caption{Comparison of (a) existing training-free video token compression methods, which increasingly rely on complex, static, hand-crafted heuristics; (b) existing learnable methods, which typically introduce trainable modules to compress tokens using fixed rules and subsequently refine their representations; and (c) our proposed learnable prototype-guided adaptive clustering method, which learns soft assignments to linearly combine original visual features while preserving consistency with their original feature-space distribution.}
\label{fig:fig1}
\end{figure*}

Most existing video token compression methods follow a training-free paradigm~\cite{hyun2025multi,tao2025dycoke,yang2025visionzip,shen2025fastvid,shao2025holitom,wang2026earlytom,li2026token,fan2026flashvid}, relying on static, hand-crafted heuristics to identify and remove or merge redundant visual tokens, as illustrated in Figure~\ref{fig:fig1}(a). Specifically, these methods adopt various strategies, including saliency- and attention-based token selection~\cite{yang2025visionzip}, similarity-based temporal token merging~\cite{tao2025dycoke,hyun2025multi}, anchor-guided aggregation~\cite{li2026token}, early-stage token compression~\cite{wang2026earlytom}, and joint spatiotemporal pruning and merging~\cite{shen2025fastvid,shao2025holitom,fan2026flashvid}. However, these methods rely on hand-crafted heuristics that are not explicitly guided by downstream task objectives, limiting their ability to preserve task-relevant visual information.

Relatively few learnable video token compression methods have been explored~\cite{liu2025hybrid,zhang2025one,yang2025pvc,wu2026marc,guo2026token}. These methods introduce trainable components at different stages of the compression pipeline, including instruction-conditioned attention and learnable global tokens~\cite{liu2025hybrid}, gated temporal attention~\cite{yang2025pvc}, supervised layer-wise token compression~\cite{zhang2025one}, reinforcement learning-based distillation~\cite{wu2026marc}, and LoRA-based adaptation with knowledge distillation~\cite{guo2026token}. As illustrated in Figure~\ref{fig:fig1}(b), despite introducing learnable parameters, these methods often rely on predefined token compression rules, limiting their flexibility in learning token assignments. Moreover, some learnable methods~\cite{ye2025voco,wang2026points} transform visual token representations during compression, requiring additional training to align the compressed representations with the LLM.

To address these limitations, we propose \textbf{COMPACT}, a novel learnable framework for video token \textbf{COM}pression via \textbf{P}rototype-guided \textbf{A}daptive \textbf{C}lus\textbf{T}ering, as illustrated in Figure~\ref{fig:fig1}(c). Specifically, COMPACT first constructs representative visual prototypes through position-aware cross-frame clustering, incorporating spatiotemporal information to capture representative visual patterns across the entire video. These initial prototypes are subsequently refined through attention-based interactions among prototypes and with the full sequence of dense visual tokens, enabling them to capture inter-prototype relationships and global visual context. Based on the refined prototypes, COMPACT combines initial token-to-prototype similarities with learned affinities to obtain differentiable soft assignment weights, which guide the weighted aggregation of original dense visual tokens into compact representations. By directly aggregating original visual features, COMPACT preserves the original visual feature space without requiring additional alignment training. Moreover, its query-agnostic design allows compressed visual tokens to be computed once and reused across multiple queries, facilitating efficient multi-turn inference. Extensive experiments on five video understanding benchmarks and three Video LLMs demonstrate that COMPACT consistently outperforms existing methods across various visual token retention ratios.

Our main contributions are summarized as follows:
\begin{itemize}
\item We propose COMPACT, a novel learnable framework for video token compression via prototype-guided adaptive clustering. By restricting the learning objective to adaptive weighted combinations of the original tokens, COMPACT preserves the original visual feature space, eliminating the need for costly additional alignment training.
\item We design a prototype-guided differentiable aggregation mechanism that initializes prototypes through position-aware cross-frame clustering, refines them via attention-based interactions, and combines initial similarities with learned affinities to produce task-adaptive soft assignments for token aggregation.
\item Extensive experiments on five video understanding benchmarks and three Video LLMs demonstrate that COMPACT consistently achieves state-of-the-art performance across various visual token retention ratios, significantly outperforming existing methods, particularly under extremely aggressive compression.
\end{itemize}
\section{Related Work}
\label{sec:related_work}

\noindent \textbf{Video Large Language Models.}
Recent Video LLMs~\cite{li2025llava,zhang2025llava,bai2025qwen2,bai2025qwen3,zhu2025internvl3,yan2026internvideo3,zhang2026thinking} extend multimodal large language models (MLLMs)~\cite{liu2024improved,tong2024cambrian,wang2024cogvlm,luo2025mono,shi2025eagle,deitke2025molmo} from static images to temporally rich video inputs, enabling open-ended video understanding and reasoning. These models typically integrate a visual encoder, a multimodal projector, and a large language model to process video frames and textual queries. Representative models include LLaVA-OneVision~\citep{li2025llava}, which unifies image and video understanding; LLaVA-Video~\citep{zhang2025llava}, which leverages large-scale synthetic video instruction data; and InternVL3~\citep{zhu2025internvl3}, which employs native multimodal pretraining. Despite their impressive capabilities, processing videos involves substantially more visual tokens than processing images, leading to higher computational and memory costs. Consequently, reducing redundant visual tokens while preserving video understanding performance has become an important research direction for improving the efficiency of Video LLMs.

\noindent \textbf{Video Token Compression.}
Most existing video token compression methods follow a training-free paradigm~\cite{hyun2025multi,tao2025dycoke,yang2025visionzip,shen2025fastvid,shao2025holitom,wang2026earlytom,li2026token,fan2026flashvid,wu2026metom,zhou2026gtr,chen2026streamingtom,cho2026floc,zhang2026rethinking}, requiring no additional training or learnable parameters and enabling plug-and-play deployment. These methods typically rely on hand-crafted heuristics for token pruning or merging. For example, EarlyTom~\cite{wang2026earlytom} merges redundant frames based on predefined similarity thresholds and applies global and local-window Top-K selection to dynamic and static frames, respectively, while AOT~\cite{li2026token} constructs attention-guided local and global anchors and aggregates tokens through predefined optimal transport operations. However, their reliance on predefined rules limits their adaptability to diverse video content.
Relatively few learnable video token compression methods have been explored~\cite{liu2025hybrid,zhang2025one,yang2025pvc,wu2026marc,guo2026token}. These methods incorporate trainable components into the compression pipeline. For instance, HICom~\cite{liu2025hybrid} uses instruction-conditioned attention and learnable global tokens to compress predefined spatiotemporal groups, while TBD~\cite{guo2026token} combines fixed FlashVID-based compression~\cite{fan2026flashvid} with LoRA adaptation~\cite{hu2021lora} and knowledge distillation. Nevertheless, such approaches still rely on predefined grouping or compression rules, limiting their flexibility in learning token assignments. Moreover, some learnable methods~\cite{ye2025voco,wang2026points} transform visual representations during compression and require additional training to adapt the compressed representations to the LLM.
In contrast, our method learns adaptive weighted combinations of the original visual tokens, enabling flexible token assignments while preserving the original visual feature space without additional alignment training. Furthermore, its query-agnostic design allows compressed tokens to be computed once and reused across multiple queries, improving multi-turn inference efficiency.
\begin{figure*}[t]
\centering
\includegraphics[width=0.95\textwidth]{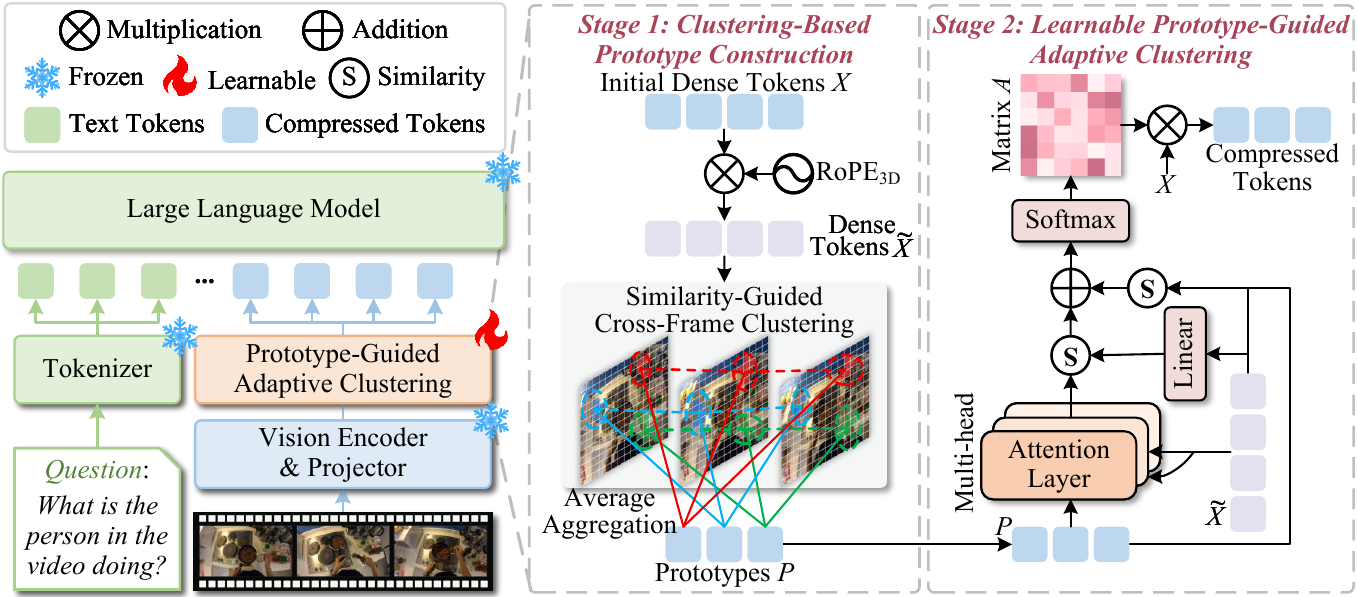}
\caption{\textbf{Overview of COMPACT}. COMPACT first constructs initial prototypes through position-aware cross-frame clustering, then refines them using learnable attention modules, followed by differentiable token-to-prototype assignment for visual token aggregation.}
\label{fig:overview}
\end{figure*}

\section{Method}
\label{sec:method}

In this section, we introduce COMPACT, a novel learnable framework for video token compression via prototype-guided adaptive clustering. The overall pipeline is illustrated in Figure~\ref{fig:overview}, with details provided in the following subsections.

\subsection{Clustering-Based Prototype Construction}

We first employ position-aware, similarity-guided cross-frame clustering to initialize a compact set of prototype tokens. Let $X\in\mathbb{R}^{N\times D}$ denote the dense visual tokens extracted by the vision encoder, where $N$ and $D$ represent the number of tokens and the feature dimension, respectively. Given a retention ratio $r$, we construct $K=\operatorname{round}(rN)$ initial prototypes, providing a content-driven initialization for subsequent learnable refinement.

\noindent \textbf{Spatiotemporal Positional Encoding.}
Visual similarity alone overlooks the spatiotemporal relationships among vision tokens, potentially leading to suboptimal clustering results. To address this limitation, we incorporate spatiotemporal information into the clustering process. Specifically, we associate each visual token $x_i$ with its spatiotemporal coordinates $(t_i,h_i,w_i)$, where $t_i$ denotes the frame index and $(h_i,w_i)$ represents its spatial position within the frame. We then apply 3D Rotary Position Embedding (RoPE)~\cite{su2024roformer}:
\begin{equation}
\widetilde{x}_i
= \operatorname{RoPE}_{3D}
\left(x_i,t_i,h_i,w_i\right).
\end{equation}
Here, $\operatorname{RoPE}_{3D}$ applies rotary transformations along the temporal, height, and width dimensions, yielding position-aware features $\widetilde{X}\in\mathbb{R}^{N\times D}$. Unlike backbone-level positional encoding (e.g., the interleaved M-RoPE in Qwen3-VL~\cite{bai2025qwen3}), our 3D RoPE is used only for clustering and leaves the backbone's positional encoding unchanged.

\noindent \textbf{Similarity-Guided Cross-Frame Clustering.}
To reduce cross-frame redundancy, we jointly cluster position-aware dense visual tokens from all sampled frames into $K$ representative clusters. Specifically, we initialize $K$ cluster centroids and iteratively assign each token to its most similar centroid based on cosine similarity, while updating each centroid as the mean of its assigned tokens. The token assignment is formulated as:
\begin{equation}
c_i = \arg\max_{k\in\{1,\ldots,K\}} \frac{\widetilde{x}_i^\top \mu_k} {\|\widetilde{x}_i\|_2\|\mu_k\|_2},
\end{equation}
where $\mu_k$ denotes the centroid of the $k$-th cluster, and $c_i$ represents the cluster assignment of the $i$-th token.

Based on the clustering results, we initialize each prototype token by averaging the position-aware visual tokens within its corresponding cluster:
\begin{equation}
p_k = \frac{1}{|\mathcal{C}_k|} \sum_{i\in\mathcal{C}_k}\widetilde{x}_i, \quad \mathcal{C}_k=\{i\mid c_i=k\},
\end{equation}
where $\mathcal{C}_k$ denotes the set of tokens assigned to the $k$-th cluster. 
For each prototype, we assign it a representative spatiotemporal position, defined by the coordinates of the earliest token in its corresponding cluster. Temporal indices take precedence, while spatial coordinates are used to resolve ties.
The resulting prototypes $P\in\mathbb{R}^{K\times D}$ are then ordered by these positions and serve as initialization for subsequent prototype refinement.

\subsection{Learnable Prototype Refinement}
Although the initial prototypes capture feature similarities, their reliance on fixed similarity-based clustering limits their adaptability to diverse video content. To address this, we introduce a lightweight refinement module that updates the prototypes through interactions among themselves and with the full visual sequence.

We first project the position-aware dense visual tokens $\widetilde{X}$ and initial prototypes $P$ from the $D$-dimensional space into a shared $d$-dimensional feature space:
\begin{equation}
Z = \operatorname{Linear}(\operatorname{LN}(\widetilde{X})), \quad R = \operatorname{Linear}(\operatorname{LN}(P)),
\end{equation}
where $\operatorname{LN}$ denotes layer normalization.

The projected prototypes are first updated through self-attention (SA)~\cite{vaswani2017attention}, followed by a residual connection~\cite{he2016deep}:
\begin{equation}
R' = R + \operatorname{SA}(\operatorname{LN}(R)).
\end{equation}
This allows each prototype to incorporate information from other prototypes. We then apply cross-attention (CA)~\cite{vaswani2017attention} with a residual connection, using the updated prototypes as queries and the projected dense visual tokens as keys and values to incorporate global visual context:
\begin{equation}
R'' = R' + \operatorname{CA}(\operatorname{LN}(R'), \operatorname{LN}(Z)).
\end{equation}
Finally, a position-wise feed-forward network (FFN) further refines the prototypes:
\begin{equation}
\widetilde{R} = R'' + \operatorname{FFN}(\operatorname{LN}(R'')).
\end{equation}

\subsection{Prototype-Guided Adaptive Clustering}

Given the position-aware dense visual tokens $\widetilde{X}$ and refined prototypes $\widetilde{R}$, we further refine the initial clustering by incorporating learned token-prototype affinities. We first compute the cosine similarity between each position-aware token $\widetilde{x}_i$ and initial prototype $p_k$, which provides a principled initialization for the token-to-prototype affinities.  We then compute the inner product between the projected token feature $z_i$ and refined prototype $\widetilde{r}_k$, incorporating the contextual information learned through prototype refinement. By combining these two terms, we obtain the token-to-prototype score matrix $S=[s_{ik}]\in\mathbb{R}^{N\times K}$, where $s_{ik}\in\mathbb{R}$ denotes the assignment score between the $i$-th token and the $k$-th prototype.

Based on the score matrix $S$, we perform adaptive clustering through differentiable soft assignments. Specifically, We apply softmax normalization along the prototype dimension to obtain the assignment matrix $A\in\mathbb{R}^{N\times K}$:
\begin{equation}
A_{ik} = \frac{\exp(s_{ik})}{\sum_{j=1}^{K}\exp(s_{ij})}, \quad \sum_{k=1}^{K}A_{ik}=1.
\end{equation}
The resulting assignment weights are then used to aggregate the original dense visual tokens into compact representations:
\begin{equation}
y_k = \frac{\sum_{i=1}^{N}A_{ik}x_i}{\sum_{i=1}^{N}A_{ik}}.
\end{equation}
Since each compressed token is a convex combination of the original visual features, the resulting representations remain in the original visual feature space, eliminating the need for costly additional alignment training. Meanwhile, the differentiable soft assignments allow gradients to propagate through the aggregation process, enabling end-to-end optimization under downstream task supervision. Furthermore, the query-agnostic nature of the clustering process allows compressed visual tokens to be computed once and reused across multiple queries, facilitating efficient multi-turn inference.

\subsection{Training Pipeline}
We freeze the vision encoder, multimodal projector, and large language model, and optimize only the attention modules and linear projection layers using the standard cross-entropy loss. This design keeps training lightweight while preserving the original visual feature-space distribution. 

The overall training pipeline consists of two stages. In the first stage, we pretrain the network using 100K multi-source video-caption pairs from LLaVA-Video-178K~\cite{zhang2025llava} to encourage the preservation of globally relevant semantic information during compression. In the second stage, we further train the network using 111K video question-answering pairs collected from LLaVA-Video-178K~\cite{zhang2025llava}, NExT-QA~\cite{xiao2021next}, CLEVRER~\cite{yi2020clevrer}, STAR~\cite{wu2024star}, and PerceptionTest~\cite{patraucean2023perception}, aiming to align the learned compression mechanism with downstream reasoning tasks and capture fine-grained visual details. Details of the data construction are provided in the supplementary material. At each stage, we randomly divide the samples into six disjoint groups, each assigned a different token retention ratio from $\{1\%, 5\%, 10\%, 15\%, 20\%, 25\%\}$, while maintaining approximately equal numbers of samples across groups.

\subsection{Discussion}
In this subsection, we discuss the connections between COMPACT and representative video token compression methods from the perspective of token aggregation.  

\noindent \textbf{General Formulation.}
Given dense visual tokens $X=\{x_i\}_{i=1}^{N}$ and a target of $K$ compressed tokens, we formulate token compression as a general weighted aggregation operation with a weight matrix $\Omega\in\mathbb{R}^{N\times K}$, where $\Omega_{ik}$ represents the contribution of the $i$-th input token to the $k$-th output token. The compressed tokens are computed as:
\begin{equation}
y_k = \frac{\sum_{i=1}^{N}\Omega_{ik}x_i}{\sum_{i=1}^{N}\Omega_{ik}}.
\end{equation}
Under this formulation, representative methods differ primarily in how their weights are determined.  
We next show that these methods can be viewed as specific instances of this general weighted aggregation framework.

\noindent \textbf{VisionZip.}
VisionZip~\cite{yang2025visionzip} selects dominant tokens based on visual attention and merges the remaining tokens into their most similar targets through hard assignments. Let $\mathcal{C}_k$ denote the group associated with the $k$-th output token. Its aggregation weights can be expressed as:
\begin{equation}
\Omega_{ik}^{\mathrm{VisionZip}} =
\begin{cases}
1, & i\in\mathcal{C}_k,\\
0, & \text{otherwise}.
\end{cases}
\end{equation}
Thus, VisionZip represents a special case of our general weighted aggregation formulation, with binary aggregation weights determined by hard token assignments.

\noindent \textbf{FastVID.}
FastVID~\cite{shen2025fastvid} selects representative anchors through density estimation and assigns redundant tokens to their nearest anchors based on cosine similarity. In its density-based token merging module, the aggregation weights for each group $\mathcal{C}_k$ with anchor $a_k$ are defined as:
\begin{equation}
\Omega_{ik}^{\mathrm{FastVID}} =
\begin{cases}
\beta, & i=a_k,\\
\frac{1-\beta}{|\mathcal{C}_k|-1}, & i\in\mathcal{C}_k\setminus\{a_k\},\\
0, & \text{otherwise},
\end{cases}
\end{equation}
where $\beta$ controls the anchor contribution. Thus, FastVID represents a special case with hard token assignments and predefined anchor-biased aggregation weights.

\noindent \textbf{AOT.}
AOT~\cite{li2026token} aggregates unselected tokens into attention-guided anchors via optimal transport (OT). Given an intra-frame transport plan $T^*$ and unselected token set $\mathcal{U}$, the aggregation weights are defined as:
\begin{equation}
\Omega_{ik}^{\mathrm{AOT}} =
\begin{cases}
1, & i=a_k,\\
\lambda_{\mathrm{intra}} T_{ik}^{*}, & i\in\mathcal{U},\\
0, & \text{otherwise},
\end{cases}
\end{equation}
where $a_k$ denotes the anchor index and $\lambda_{\mathrm{intra}}$ controls the aggregation strength. Thus, AOT represents a special case with transport-based aggregation weights determined by the optimal transport plan.

\noindent \textbf{EarlyTom.}
EarlyTom~\cite{wang2026earlytom} reduces temporal redundancy through similarity-guided merging of adjacent frames within the vision encoder. Let $s_t$ and $s_{t+1}$ denote their similarity scores, and $\pi(t,p_k)$ denote the index of the token at spatial position $p_k$ in frame $t$. The aggregation weights are defined as:
\begin{equation}
\Omega_{ik}^{\mathrm{EarlyTom}} =
\begin{cases}
s_t, & i=\pi(t,p_k),\\
s_{t+1}, & i=\pi(t+1,p_k),\\
0, & \text{otherwise}.
\end{cases}
\end{equation}
Thus, EarlyTom represents a special case of weighted aggregation, with nonzero weights restricted to spatially aligned tokens in adjacent frames.

Overall, these methods rely on different hand-crafted aggregation strategies, including hard similarity-based grouping, anchor-biased weighting, optimal transport, and similarity-weighted temporal fusion. Despite their differences, they can be viewed as special cases of our general weighted aggregation formulation, with specific constraints or rules for determining aggregation weights. In contrast, COMPACT learns flexible token-to-prototype aggregation weights through differentiable soft assignments under downstream task supervision, enabling more adaptive and task-aware visual token compression.
\section{Experiments}
\label{sec:experiments}
\subsection{Experimental Settings}

\noindent \textbf{Benchmarks and Compared Baselines.}
We evaluate our method on five widely used video understanding benchmarks: VideoMME~\citep{fu2025video}, LongVideoBench~\citep{wu2024longvideobench}, MLVU~\cite{zhou2025mlvu}, MVBench~\citep{li2024mvbench}, and EgoSchema~\citep{mangalam2023egoschema}. These benchmarks cover videos of varying durations and scene complexities, enabling a comprehensive assessment of our method's effectiveness and generalizability.
We compare our method with several recent representative video token compression methods, including FastVID~\cite{shen2025fastvid}, VisionZip~\cite{yang2025visionzip}, EarlyTom~\cite{wang2026earlytom}, and AOT~\cite{li2026token}.

\noindent \textbf{Implementation Details.}
For both training stages, we use a global batch size of 128 and a learning rate of $2\times10^{-5}$. We project the position-aware dense visual tokens and initial prototypes into a shared feature space with dimension $d=256$ and employ multi-head attention with 8 heads. During prototype construction, we limit the clustering process to at most 10 iterations. We evaluate our method on three representative Video LLMs, namely LLaVA-OneVision-7B~\cite{li2025llava}, LLaVA-Video-7B~\cite{zhang2025llava}, and InternVL3-8B~\cite{zhu2025internvl3}, to assess its generalizability across diverse architectures. Following the official evaluation settings, LLaVA-OneVision-7B, LLaVA-Video-7B, and InternVL3-8B uniformly sample 32, 64, and 32 frames, respectively. All evaluations are conducted on NVIDIA A800 80GB GPUs using LMMs-Eval~\citep{zhang2025lmms}.

\begin{table*}[t]
\caption{Comparison with state-of-the-art methods on LLaVA-OneVision-7B. ``Avg. Score'' and ``Avg. \%'' denote the average score across benchmarks and the relative performance to the unpruned model. FastVID and AOT are omitted at 1\% because their structural constraints require more tokens than the available budget. The best and second-best results are in \textbf{bold} and \underline{underlined}, respectively.}
\label{tab:llava_ov}
\centering
\resizebox{\textwidth}{!}{%
\begin{tabular}{l|c|c|ccccc|cc}
\toprule
\multirow{2}{*}{Method}
& \multirow{2}{*}{Source}
& \multirow{2}{*}{\shortstack{Before LLM\\Retained Ratio}}
& \multirow{2}{*}{\shortstack{VideoMME\\$\uparrow$}}
& \multirow{2}{*}{\shortstack{LongVideoBench\\$\uparrow$}}
& \multirow{2}{*}{\shortstack{MLVU\\$\uparrow$}}
& \multirow{2}{*}{\shortstack{MVBench\\$\uparrow$}}
& \multirow{2}{*}{\shortstack{EgoSchema\\$\uparrow$}}
& \multicolumn{2}{|c}{Avg. $\uparrow$} \\
& & & & & & & & Score & \% \\
\midrule
\rowcolor[RGB]{235,235,235}
LLaVA-OV-7B~\cite{li2025llava} & -- & 100\% & 58.6 & 56.4 & 63.0 & 58.3 & 60.4 & 59.3 & 100.0 \\
\midrule
FastVID~\cite{shen2025fastvid} & NeurIPS'25 & 15\% & \underline{57.7} & \underline{56.2} & 61.0 & 56.0 & 57.4 & 57.7 & 97.2 \\
VisionZip~\cite{yang2025visionzip} & CVPR'25 & 15\% & 56.1 & 54.4 & 60.8 & 56.5 & 59.8 & 57.5 & 96.9 \\
EarlyTom~\cite{wang2026earlytom} & CVPR'26 & 15\% & 56.9 & 54.4 & \underline{61.2} & 57.5 & 60.2 & 58.0 & 97.8 \\
AOT~\cite{li2026token} & CVPR'26 & 15\% & 56.6 & 55.2 & 60.1 & \textbf{57.8} & \textbf{61.3} & \underline{58.2} & \underline{98.1} \\
\rowcolor[RGB]{234,234,255}
\textbf{COMPACT} & -- & 15\% & \textbf{58.7} & \textbf{58.3} & \textbf{63.1} & \underline{57.6} & \underline{60.5} & \textbf{59.7} & \textbf{100.5} \\
\midrule
FastVID~\cite{shen2025fastvid} & NeurIPS'25 & 10\% & \textbf{57.3} & \underline{56.3} & \underline{61.1} & 55.9 & 56.5 & \underline{57.4} & \underline{96.7} \\
VisionZip~\cite{yang2025visionzip} & CVPR'25 & 10\% & 53.4 & 49.3 & 58.9 & 53.5 & 58.0 & 54.6 & 92.0 \\
EarlyTom~\cite{wang2026earlytom} & CVPR'26 & 10\% & 55.8 & 52.4 & 59.8 & 56.5 & \underline{60.1} & 56.9 & 95.9 \\
AOT~\cite{li2026token} & CVPR'26 & 10\% & 56.1 & 54.2 & 58.2 & \underline{57.0} & \textbf{60.6} & 57.2 & 96.4 \\
\rowcolor[RGB]{234,234,255}
\textbf{COMPACT} & -- & 10\% & \underline{57.1} & \textbf{56.6} & \textbf{61.6} & \textbf{57.4} & \underline{60.1} & \textbf{58.6} & \textbf{98.7} \\
\midrule
FastVID~\cite{shen2025fastvid} & NeurIPS'25 & 5\% & \underline{53.8} & \underline{51.7} & \underline{57.6} & 53.8 & 56.9 & \underline{54.8} & \underline{92.3} \\
VisionZip~\cite{yang2025visionzip} & CVPR'25 & 5\% & 48.6 & 46.5 & 54.1 & 45.3 & 51.8 & 49.3 & 83.0 \\
EarlyTom~\cite{wang2026earlytom} & CVPR'26 & 5\% & 51.4 & 50.2 & 57.2 & 51.9 & 58.2 & 53.8 & 90.6 \\
AOT~\cite{li2026token} & CVPR'26 & 5\% & 52.3 & 50.5 & 56.8 & \underline{54.7} & \textbf{59.5} & \underline{54.8} & \underline{92.3} \\
\rowcolor[RGB]{234,234,255}
\textbf{COMPACT} & -- & 5\% & \textbf{56.6} & \textbf{55.9} & \textbf{61.1} & \textbf{56.5} & \underline{58.7} & \textbf{57.7} & \textbf{97.3} \\
\midrule
VisionZip~\cite{yang2025visionzip} & CVPR'25 & 1\% & 44.4 & 44.7 & \underline{51.7} & 40.6 & 43.7 & 45.0 & 75.9 \\
EarlyTom~\cite{wang2026earlytom} & CVPR'26 & 1\% & \underline{45.9} & \underline{45.3} & 50.9 & \underline{43.3} & \underline{48.1} & \underline{46.7} & \underline{78.7} \\
\rowcolor[RGB]{234,234,255}
\textbf{COMPACT} & -- & 1\% & \textbf{52.2} & \textbf{50.6} & \textbf{57.2} & \textbf{52.6} & \textbf{55.1} & \textbf{53.5} & \textbf{90.2} \\
\bottomrule
\end{tabular}%
}
\end{table*}

\begin{table*}[t]
\caption{Comparison with state-of-the-art methods on LLaVA-Video-7B. ``Avg. Score'' and ``Avg. \%'' denote the average score across benchmarks and the relative performance to the unpruned model. AOT is omitted at 1\% because its structural constraints require more tokens than the available budget. The best and second-best results are in \textbf{bold} and \underline{underlined}, respectively.}
\label{tab:llava_video}
\centering
\resizebox{\textwidth}{!}{%
\begin{tabular}{l|c|c|ccccc|cc}
\toprule
\multirow{2}{*}{Method}
& \multirow{2}{*}{Source}
& \multirow{2}{*}{\shortstack{Before LLM\\Retained Ratio}}
& \multirow{2}{*}{\shortstack{VideoMME\\$\uparrow$}}
& \multirow{2}{*}{\shortstack{LongVideoBench\\$\uparrow$}}
& \multirow{2}{*}{\shortstack{MLVU\\$\uparrow$}}
& \multirow{2}{*}{\shortstack{MVBench\\$\uparrow$}}
& \multirow{2}{*}{\shortstack{EgoSchema\\$\uparrow$}}
& \multicolumn{2}{|c}{Avg. $\uparrow$} \\
& & & & & & & & Score & \% \\
\midrule
\rowcolor[RGB]{235,235,235}
LLaVA-Video-7B~\cite{zhang2025llava}
& -- & 100\% & 64.3 & 58.9 & 68.6 & 60.4 & 57.2 & 61.9 & 100.0 \\
\midrule
VisionZip~\cite{yang2025visionzip}
& CVPR'25 & 15\%
& 60.7 & 54.7 & \underline{64.6} & 56.7
& \underline{54.7} & 58.3 & 94.2 \\
EarlyTom~\cite{wang2026earlytom}
& CVPR'26 & 15\%
& \underline{61.3} & 53.9 & 62.0 & 55.8
& \underline{54.7} & 57.5 & 93.0 \\
AOT~\cite{li2026token}
& CVPR'26 & 15\%
& \textbf{62.0} & \underline{55.0} & 63.7
& \underline{57.8} & \textbf{55.2}
& \underline{58.7} & \underline{94.9} \\
\rowcolor[RGB]{234,234,255}
\textbf{COMPACT}
& -- & 15\%
& 61.1 & \textbf{57.4} & \textbf{66.1}
& \textbf{60.6} & 53.1
& \textbf{59.7} & \textbf{96.4} \\
\midrule
VisionZip~\cite{yang2025visionzip}
& CVPR'25 & 10\%
& 59.0 & 53.9 & \underline{62.8} & 57.8
& \underline{53.1} & 57.3 & 92.7 \\
EarlyTom~\cite{wang2026earlytom}
& CVPR'26 & 10\%
& 56.0 & 50.9 & 57.0 & 56.8
& 47.0 & 53.5 & 86.6 \\
AOT~\cite{li2026token}
& CVPR'26 & 10\%
& \textbf{60.8} & \underline{54.1} & 62.5
& \underline{59.1} & \textbf{53.4}
& \underline{58.0} & \underline{93.7} \\
\rowcolor[RGB]{234,234,255}
\textbf{COMPACT}
& -- & 10\%
& \underline{60.6} & \textbf{57.5} & \textbf{65.4}
& \textbf{60.1} & 52.0
& \textbf{59.1} & \textbf{95.5} \\
\midrule
VisionZip~\cite{yang2025visionzip}
& CVPR'25 & 5\%
& 53.6 & \underline{50.1} & 56.0 & 49.9
& \underline{44.9} & 50.9 & 82.3 \\
EarlyTom~\cite{wang2026earlytom}
& CVPR'26 & 5\%
& 53.7 & 47.9 & 55.3 & 48.1
& 44.7 & 49.9 & 80.7 \\
AOT~\cite{li2026token}
& CVPR'26 & 5\%
& \underline{56.8} & 49.4 & \underline{59.0}
& \underline{52.6} & \textbf{50.1}
& \underline{53.6} & \underline{86.6} \\
\rowcolor[RGB]{234,234,255}
\textbf{COMPACT}
& -- & 5\%
& \textbf{59.7} & \textbf{54.3} & \textbf{63.7}
& \textbf{59.3} & \textbf{50.1}
& \textbf{57.4} & \textbf{92.8} \\
\midrule
VisionZip~\cite{yang2025visionzip}
& CVPR'25 & 1\%
& \underline{46.6} & \underline{45.0}
& \underline{50.8} & \underline{43.4}
& \underline{36.5} & \underline{44.5}
& \underline{71.8} \\
EarlyTom~\cite{wang2026earlytom}
& CVPR'26 & 1\%
& 44.6 & 43.4 & 48.4 & 41.8
& 33.0 & 42.2 & 68.3 \\
\rowcolor[RGB]{234,234,255}
\textbf{COMPACT}
& -- & 1\%
& \textbf{56.4} & \textbf{51.8} & \textbf{58.1}
& \textbf{56.2} & \textbf{45.8}
& \textbf{53.7} & \textbf{86.7} \\
\bottomrule
\end{tabular}%
}
\end{table*}
\begin{table*}[!ht]
\caption{Comparison with state-of-the-art methods on InternVL3-8B. ``Avg. Score'' and ``Avg. \%'' denote the average score across benchmarks and the relative performance to the unpruned model. AOT is omitted at 1\% because its structural constraints require more tokens than the available budget. The best and second-best results are in \textbf{bold} and \underline{underlined}, respectively.}
\label{tab:internvl3_8b}
\centering
\resizebox{\textwidth}{!}{%
\begin{tabular}{l|c|c|ccccc|cc}
\toprule
\multirow{2}{*}{Method}
& \multirow{2}{*}{Source}
& \multirow{2}{*}{\shortstack{Before LLM\\Retained Ratio}}
& \multirow{2}{*}{\shortstack{VideoMME\\$\uparrow$}}
& \multirow{2}{*}{\shortstack{LongVideoBench\\$\uparrow$}}
& \multirow{2}{*}{\shortstack{MLVU\\$\uparrow$}}
& \multirow{2}{*}{\shortstack{MVBench\\$\uparrow$}}
& \multirow{2}{*}{\shortstack{EgoSchema\\$\uparrow$}}
& \multicolumn{2}{|c}{Avg. $\uparrow$} \\
& & & & & & & & Score & \% \\
\midrule
\rowcolor[RGB]{235,235,235}
InternVL3-8B~\cite{zhu2025internvl3} & -- & 100\% & 64.9 & 58.7 & 68.8 & 74.1 & 62.6 & 65.8 & 100.0 \\
\midrule
VisionZip~\cite{yang2025visionzip} & CVPR'25 & 15\% & \underline{63.3} & \textbf{57.4} & \underline{67.8} & 72.1 & \underline{60.0} & \underline{64.1} & \underline{97.4} \\
EarlyTom~\cite{wang2026earlytom} & CVPR'26 & 15\% & 62.0 & \textbf{57.4} & 67.6 & \underline{72.8} & \textbf{60.1} & 64.0 & 97.2 \\
AOT~\cite{li2026token} & CVPR'26 & 15\% & 60.1 & \underline{54.4} & 63.9 & 71.8 & 57.4 & 61.5 & 93.5 \\
\rowcolor[RGB]{234,234,255}
\textbf{COMPACT} & -- & 15\% & \textbf{63.7} & \textbf{57.4} & \textbf{68.9} & \textbf{72.9} & \underline{60.0} & \textbf{64.6} & \textbf{98.1} \\
\midrule
VisionZip~\cite{yang2025visionzip} & CVPR'25 & 10\% & \underline{62.3} & 55.2 & \underline{66.1} & 68.2 & \underline{58.6} & \underline{62.1} & \underline{94.3} \\
EarlyTom~\cite{wang2026earlytom} & CVPR'26 & 10\% & 60.3 & \underline{55.7} & 63.9 & \underline{70.6} & 58.1 & 61.7 & 93.8 \\
AOT~\cite{li2026token} & CVPR'26 & 10\% & 59.7 & 54.0 & 62.7 & 69.2 & 55.0 & 60.1 & 91.3 \\
\rowcolor[RGB]{234,234,255}
\textbf{COMPACT} & -- & 10\% & \textbf{63.5} & \textbf{56.5} & \textbf{67.8} & \textbf{72.4} & \textbf{59.1} & \textbf{63.9} & \textbf{97.1} \\
\midrule
VisionZip~\cite{yang2025visionzip} & CVPR'25 & 5\% & \underline{59.3} & 52.0 & \underline{62.9} & 61.7 & \underline{55.8} & 58.3 & 88.6 \\
EarlyTom~\cite{wang2026earlytom} & CVPR'26 & 5\% & 57.2 & \underline{53.8} & 62.8 & \underline{66.1} & 55.5 & \underline{59.1} & \underline{89.8} \\
AOT~\cite{li2026token} & CVPR'26 & 5\% & 56.6 & 51.9 & 60.2 & 65.3 & 50.9 & 57.0 & 86.6 \\
\rowcolor[RGB]{234,234,255}
\textbf{COMPACT} & -- & 5\% & \textbf{62.5} & \textbf{55.6} & \textbf{67.3} & \textbf{69.8} & \textbf{57.0} & \textbf{62.4} & \textbf{94.8} \\
\midrule
VisionZip~\cite{yang2025visionzip} & CVPR'25 & 1\% & 50.3 & 46.6 & 53.0 & 46.1 & 44.1 & 48.0 & 73.0 \\
EarlyTom~\cite{wang2026earlytom} & CVPR'26 & 1\% & \underline{53.0} & \underline{49.2} & \underline{55.3} & \underline{51.7} & \underline{44.3} & \underline{50.7} & \underline{77.0} \\
\rowcolor[RGB]{234,234,255}
\textbf{COMPACT} & -- & 1\% & \textbf{58.9} & \textbf{54.1} & \textbf{63.2} & \textbf{64.0} & \textbf{51.6} & \textbf{58.3} & \textbf{88.7} \\
\bottomrule
\end{tabular}%
}
\end{table*}

\subsection{Main Results}

We compare COMPACT with state-of-the-art baselines on three representative Video LLMs with different architectures across six visual token retention ratios (1\%, 5\%, 10\%, 15\%, 20\%, and 25\%). The results at 1\%, 5\%, 10\%, and 15\% retention are reported in Tables~\ref{tab:llava_ov}, \ref{tab:llava_video}, and \ref{tab:internvl3_8b}, while the results at 20\% and 25\% retention are provided in the supplementary material.

As shown in the tables, COMPACT consistently and substantially outperforms all competing methods in terms of average scores across the evaluated retention ratios on all three backbones. Specifically, on LLaVA-OneVision-7B, COMPACT achieves average scores of 59.7, 58.6, 57.7, and 53.5 at 15\%, 10\%, 5\%, and 1\% retention, respectively. Similar improvements are observed on LLaVA-Video-7B and InternVL3-8B, demonstrating its effectiveness across different model architectures. Notably, the performance gains become more pronounced under aggressive compression. At just 1\% retention, COMPACT outperforms the strongest competing baselines by 6.8, 9.2, and 7.6 percentage points on the three backbones, respectively. These results highlight the advantages of prototype-guided adaptive clustering, particularly under extremely low visual token budgets.

\begin{figure}[t]
\centering
\includegraphics[width=\columnwidth]{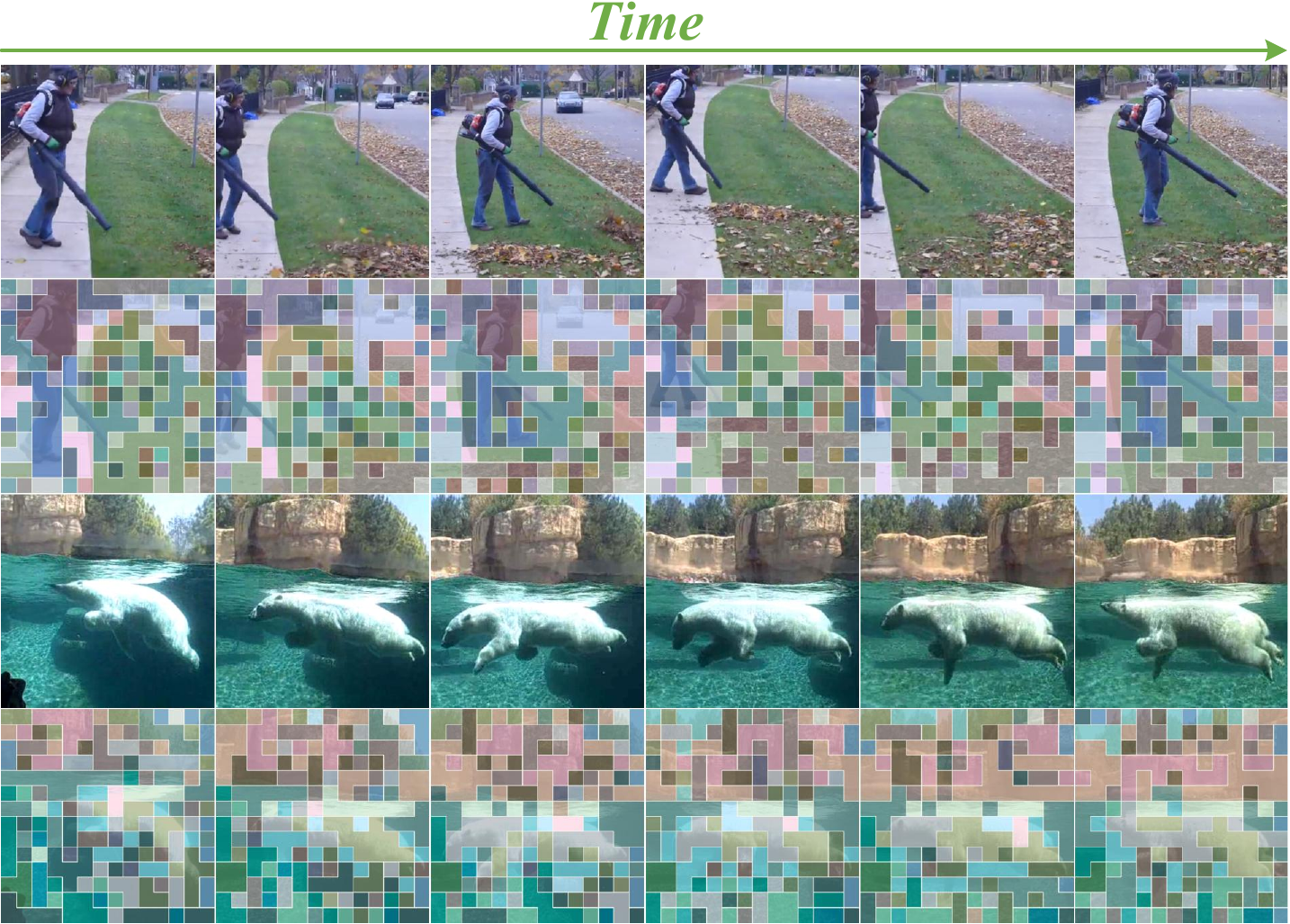}
\caption{Cross-frame video clustering visualizations produced by COMPACT. The same color indicates the same cluster.}
\label{fig:vis}
\end{figure}

Figure~\ref{fig:vis} further illustrates the cross-frame clustering results of COMPACT, where tokens assigned to the same cluster are shown in the same color. The visualization shows that the learned prototypes capture representative visual patterns across frames, helping preserve informative visual content under limited token budgets.

\subsection{Ablation Studies}
We conduct comprehensive ablation studies on LLaVA-OneVision-7B at a 15\% visual token retention ratio.

\noindent \textbf{Effect of Spatiotemporal Positional Encoding.}
We first examine the effect of 3D RoPE in the initial clustering stage. As shown in Table~\ref{tab:ablation}, removing 3D RoPE consistently degrades performance across all five benchmarks. This indicates that incorporating spatiotemporal positional information helps distinguish visually similar tokens at different spatial and temporal positions, leading to more effective prototype initialization.

\noindent \textbf{Effect of Cross-Frame Clustering.}
We further compare joint cross-frame clustering with independent per-frame clustering. As shown in Table~\ref{tab:ablation}, cross-frame clustering consistently outperforms per-frame clustering across all three benchmarks. This indicates that joint clustering better exploits temporal redundancy and captures representative visual patterns across the entire video, leading to more effective prototype construction.

\noindent \textbf{Effect of Prototype Refinement.}
We examine the contribution of learnable prototype refinement by directly using the initial prototypes for compression without further refinement. As shown in Table~\ref{tab:ablation}, the full model consistently outperforms this variant across all five benchmarks. This indicates that attention-based refinement under downstream task supervision enables prototypes to capture task-relevant interactions and incorporate global visual context.

\noindent \textbf{Effect of Soft Assignment.}
We compare hard and soft token-to-prototype assignments in the adaptive clustering stage. Under hard assignment, each token is assigned exclusively to its highest-scoring prototype, i.e., $c_i=\arg\max_k s_{ik}$, resulting in a one-hot assignment matrix $H$, where $H_{ik}=1$ if $k=c_i$ and $0$ otherwise. During training, we employ hard Gumbel-Softmax, using discrete assignments in the forward pass and a continuous relaxation for gradient estimation in the backward pass. At inference, we use deterministic argmax without Gumbel noise. The compressed tokens are obtained by averaging the original, unrotated visual features within each cluster. As shown in Table~\ref{tab:ablation}, soft assignment consistently outperforms hard assignment across all five benchmarks, demonstrating the benefits of continuous assignments for flexible feature aggregation and end-to-end optimization.

\noindent \textbf{Effect of Two-Stage Training.}
We finally examine the effect of two-stage training by comparing it with single-stage training using only the 111K video question-answering pairs. As shown in Table~\ref{tab:ablation}, the full model consistently outperforms the single-stage variant across all five benchmarks, demonstrating the benefits of video-caption pretraining for preserving global semantic information and improving downstream video understanding.

\begin{table}[t]
\caption{Ablation studies on LLaVA-OneVision-7B at a 15\% visual token retention ratio. LVB denotes LongVideoBench.}
\label{tab:ablation}
\centering
\resizebox{\columnwidth}{!}{
\begin{tabular}{l|ccccc}
\toprule
\multirow{2}{*}{Setting}
& VideoMME & LVB & MLVU & MVBench & EgoSchema \\
& $\uparrow$ & $\uparrow$ & $\uparrow$ & $\uparrow$ & $\uparrow$ \\
\midrule
w/o 3D RoPE & 57.0 & 56.1 & 60.9 & 56.7 & 60.1 \\
w/o Cross-Frame & 58.1 & 57.3 & 61.7 & 55.8 & 60.1 \\
w/o Prototype Refinement & 57.3 & 56.9 & 61.8 & 57.4 & 60.1 \\
w/o Soft Assignment & 57.6 & 57.0 & 61.8 & 57.5 & 59.8 \\
w/o Two-Stage Training & 57.6 & 57.8 & 63.0 & 57.2 & 60.2 \\
\rowcolor[RGB]{234,234,255}
\textbf{COMPACT} & \textbf{58.7} & \textbf{58.3} & \textbf{63.1} & \textbf{57.6} & \textbf{60.5} \\
\bottomrule
\end{tabular}
}
\end{table}

\subsection{Efficiency Analysis}
We evaluate the efficiency of COMPACT on LLaVA-OneVision-7B, LLaVA-Video-7B, and InternVL3-8B at a 15\% visual token retention ratio. All experiments are conducted on a single NVIDIA A800 GPU using the same video-prompt pairs. We measure the floating-point operations (FLOPs) of the vision encoder and LLM prefill stage, excluding multimodal projection and compression costs. We also report the time to first token (TTFT), covering visual encoding, projection, compression, and first-token generation, but excluding video decoding and preprocessing. TTFT is averaged over five runs. In addition, we report the end-to-end generation throughput, calculated as the total number of tokens generated across five runs divided by the corresponding total generation time.

\begin{table}[t]
\caption{Efficiency analysis of COMPACT on LLaVA-OneVision-7B at a 15\% visual token retention ratio.}
\label{tab:efficiency}
\centering
\resizebox{\columnwidth}{!}{%
\begin{tabular}{l|ccc|cc}
\toprule
\multirow{2}{*}{Method}
& \multirow{2}{*}{\shortstack{Prefilling\\FLOPs (T) $\downarrow$}}
& \multirow{2}{*}{\shortstack{TTFT\\(ms) $\downarrow$}}
& \multirow{2}{*}{\shortstack{Throughput\\(tokens/s) $\uparrow$}}
& \multicolumn{2}{c}{Avg. $\uparrow$} \\
& & & & Score & \% \\
\midrule
LLaVA-OV-7B~\cite{li2025llava}
& 125.6 & 944.5 & 30.9 & 59.3 & 100.0 \\
AOT~\cite{li2026token}
& 35.0 & 709.0 & 36.0 & 58.2 & 98.1 \\
\rowcolor[RGB]{234,234,255}
\textbf{COMPACT}
& \textbf{34.6} & \textbf{587.3} & \textbf{37.1} & \textbf{59.7} & \textbf{100.5} \\
\bottomrule
\end{tabular}%
}
\end{table}

As shown in Table~\ref{tab:efficiency}, COMPACT achieves the best overall efficiency-performance trade-off. Compared with LLaVA-OneVision-7B, COMPACT reduces prefill FLOPs from 125.6T to 34.6T and TTFT from 944.5 ms to 587.3 ms. It also achieves a generation throughput of 37.1 tokens/s, exceeding both the original model (30.9 tokens/s) and AOT (36.0 tokens/s). Notably, despite reducing visual tokens by 85\%, COMPACT achieves an average score of 59.7, even surpassing the uncompressed model (59.3). These results demonstrate that COMPACT substantially improves inference efficiency without compromising model performance. Additional results on LLaVA-Video-7B and InternVL3-8B are provided in the supplementary material.
\section{Conclusion}
\label{sec:conclusion}
In this work, we present COMPACT, a novel learnable framework for video token compression via prototype-guided adaptive clustering. By combining position-aware prototype initialization, attention-based refinement, and differentiable soft assignments, COMPACT enables task-adaptive token aggregation under downstream supervision. Importantly, it learns weighted combinations of original visual tokens, preserving the original visual feature space without costly alignment training. Its query-agnostic design also allows compressed tokens to be reused across multiple queries. Extensive experiments demonstrate that COMPACT consistently outperforms existing compression methods, particularly under extremely low token budgets.

\section*{LLM Usage Statement}
In this work, we used Large Language Models (LLMs) to improve the clarity, grammar, and readability of the paper. After using these tools, we carefully reviewed and edited the content as needed and take full responsibility for the final content of this work.

{
    \small
    \bibliographystyle{ieeenat_fullname}
    \bibliography{main}
}

\clearpage
\hypersetup{pageanchor=false}
\setcounter{page}{1}
\maketitlesupplementary

\setcounter{section}{0}
\renewcommand{\thesection}{\Alph{section}}
\renewcommand{\theHsection}{supp.\Alph{section}}

\section*{Overview}
This supplementary material provides additional evaluations, training data details, experimental analyses, implementation details, and qualitative visualizations of COMPACT. The contents are organized as follows:
\begin{itemize}
\item \textbf{Section~\ref{sec:evaluation_benchmarks}} introduces the five video understanding benchmarks used in our experiments, including VideoMME, LongVideoBench, MLVU, MVBench, and EgoSchema. We describe their key characteristics and evaluation focuses, covering diverse video content, long-context comprehension, temporal reasoning, and egocentric understanding.
\item \textbf{Section~\ref{sec:training-data-composition}} details the composition of the training data used in the two-stage training process. We describe the video captioning and video question-answering datasets, their sample distributions, and the data consistency maintained across different backbone architectures.
\item \textbf{Section~\ref{sec:more_experimental_results}} presents additional experimental results, including extended performance comparisons at different token retention ratios across three Video LLM backbones and a computational efficiency analysis comparing COMPACT with AOT in terms of clustering core time, total compression time, and LLM prefill time.
\item \textbf{Section~\ref{sec:implementation-details}} provides additional implementation details of COMPACT and the compared methods. We describe the reproduction settings and backbone-specific adaptations of existing methods, visual input processing and token budget configurations, network architecture, numerical implementation, and training hyperparameters.
\item \textbf{Section~\ref{sec:unified_assignment}} presents a unified weighted aggregation formulation for representative visual token compression methods, including token selection, similarity-based merging, and anchor-based aggregation. We show how these methods can be expressed as special cases of the general aggregation framework underlying COMPACT.
\item \textbf{Section~\ref{sec:more_visualizations}} presents additional qualitative visualizations of COMPACT, including cross-frame clustering results across diverse videos and clustering assignments at different visual token retention ratios. These visualizations illustrate its clustering behavior and the consistency of visual grouping across frames and compression levels.
\end{itemize}

\section{Evaluation Benchmarks}
\label{sec:evaluation_benchmarks}

We evaluate COMPACT on five widely adopted video understanding benchmarks, covering diverse video content and durations, long-context comprehension, fine-grained understanding, temporal reasoning, and egocentric scenarios. Together, these benchmarks provide a comprehensive evaluation across diverse video understanding tasks.

\noindent \textbf{VideoMME.}
VideoMME~\citep{fu2025video} evaluates video understanding across diverse domains and temporal scales. It contains 900 videos spanning 6 major domains and 30 fine-grained categories, with durations ranging from 11 seconds to 1 hour. It provides 2,700 human-annotated multiple-choice question-answer pairs covering video perception, comprehension, and reasoning.

\noindent \textbf{LongVideoBench.}
LongVideoBench~\citep{wu2024longvideobench} assesses long-context video-language understanding. It comprises 3,763 videos ranging from 8 seconds to 1 hour and 6,678 human-annotated multiple-choice questions. Its referring-reasoning formulation requires models to identify relevant visual and linguistic evidence in long videos and perform fine-grained reasoning across 17 perceptual and relational question categories.

\noindent \textbf{MLVU.}
MLVU~\citep{zhou2025mlvu} is a multi-task benchmark for long-video understanding, covering holistic comprehension and fine-grained detail recognition. It assesses the ability to integrate global video context and retrieve specific information across extended temporal spans, emphasizing long-range temporal reasoning.

\noindent \textbf{MVBench.}
MVBench~\citep{li2024mvbench} evaluates temporal video understanding through 20 video-centric tasks that extend static image-based tasks to dynamic scenarios. It assesses temporal perception and reasoning, including action recognition, temporal dynamics, and event relationships.

\noindent \textbf{EgoSchema.}
EgoSchema~\citep{mangalam2023egoschema} focuses on long-horizon reasoning in egocentric videos. It contains approximately 5,000 human-annotated five-option multiple-choice questions derived from 250 hours of egocentric video data. Each question is associated with a video clip of approximately 3 minutes, covering extended temporal interactions across 289 video segments. The benchmark evaluates object and action tracking as well as long-range temporal reasoning.

\section{Training Data Composition}
\label{sec:training-data-composition}

We construct the training data from two complementary sources for the caption and QA stages, combining video-level descriptions and question-answering annotations to provide diverse training supervision.

\noindent \textbf{Video Captioning Supervision.}
The first stage uses 99,163 video-caption pairs from LLaVA-Video-178K~\cite{zhang2025llava} for caption-based pretraining, encouraging the network to preserve globally relevant semantic information during compression. The videos are distributed across four duration ranges: 11,985 videos of 0--30\,s, 30,508 of 30--60\,s, 28,861 of 1--2\,min, and 27,809 of 2--3\,min.

\noindent \textbf{Video Question-Answering Supervision.}
The second-stage training data comprise 111,309 video question-answering pairs collected from five datasets: 77,737 from LLaVA-Video-178K~\cite{zhang2025llava}, 11,455 from STAR~\cite{wu2024star}, 8,220 from CLEVRER~\cite{yi2020clevrer}, 7,549 from NExT-QA~\cite{xiao2021next}, and 6,348 from PerceptionTest~\cite{patraucean2023perception}. These pairs include 104,273 multiple-choice, 4,934 free-form, and 2,102 numerical-answer samples. In terms of video duration, 48,854 pairs correspond to videos of 0--30\,s, 31,474 to 30--60\,s, 2,050 to 1--2\,min, and 28,931 to 2--3\,min.

\noindent \textbf{Training Data Consistency.}
To ensure consistency across different backbone architectures, all experiments use the same video captioning and video question-answering training data for LLaVA-OneVision-7B~\cite{li2025llava}, LLaVA-Video-7B~\cite{zhang2025llava}, and InternVL3-8B~\cite{zhu2025internvl3}.

\begin{table*}[t]
\caption{Comparison with state-of-the-art methods on LLaVA-OneVision-7B. ``Avg. Score'' and ``Avg. \%'' denote the average score across benchmarks and the relative performance to the unpruned model. The best and second-best results are in \textbf{bold} and \underline{underlined}, respectively.}
\label{tab:sup_llava_ov}
\centering
\resizebox{\textwidth}{!}{%
\begin{tabular}{l|c|c|ccccc|cc}
\toprule
\multirow{2}{*}{Method}
& \multirow{2}{*}{Source}
& \multirow{2}{*}{\shortstack{Before LLM\\Retained Ratio}}
& \multirow{2}{*}{\shortstack{VideoMME\\$\uparrow$}}
& \multirow{2}{*}{\shortstack{LongVideoBench\\$\uparrow$}}
& \multirow{2}{*}{\shortstack{MLVU\\$\uparrow$}}
& \multirow{2}{*}{\shortstack{MVBench\\$\uparrow$}}
& \multirow{2}{*}{\shortstack{EgoSchema\\$\uparrow$}}
& \multicolumn{2}{|c}{Avg. $\uparrow$} \\
& & & & & & & & Score & \% \\
\midrule
\rowcolor[RGB]{235,235,235}
LLaVA-OV-7B~\cite{li2025llava} & -- & 100\% & 58.6 & 56.4 & 63.0 & 58.3 & 60.4 & 59.3 & 100.0 \\
\midrule
FastVID~\cite{shen2025fastvid} & NeurIPS'25 & 25\% & 58.0 & 56.3 & 61.7 & 56.5 & 58.2 & 58.1 & 98.0 \\
VisionZip~\cite{yang2025visionzip} & CVPR'25 & 25\% & \underline{58.2} & \underline{56.5} & \underline{63.1} & \underline{57.9} & 60.3 & \underline{59.2} & \underline{99.8} \\
EarlyTom~\cite{wang2026earlytom} & CVPR'26 & 25\% & \textbf{58.5} & 56.3 & 62.6 & 57.4 & 60.5 & 59.1 & 99.5 \\
AOT~\cite{li2026token} & CVPR'26 & 25\% & 57.5 & 56.3 & 61.3 & \textbf{58.7} & \textbf{61.3} & 59.0 & 99.5 \\
\rowcolor[RGB]{234,234,255}
\textbf{COMPACT} & -- & 25\% & 58.1 & \textbf{57.6} & \textbf{63.3} & 57.5 & \underline{61.0} & \textbf{59.5} & \textbf{100.3} \\
\midrule
FastVID~\cite{shen2025fastvid} & NeurIPS'25 & 20\% & 57.9 & \underline{57.1} & 61.2 & 56.3 & 57.9 & 58.0 & 97.8 \\
VisionZip~\cite{yang2025visionzip} & CVPR'25 & 20\% & 57.9 & 55.2 & 62.4 & 57.7 & 59.8 & 58.6 & 98.7 \\
EarlyTom~\cite{wang2026earlytom} & CVPR'26 & 20\% & \underline{58.0} & 55.6 & \underline{62.7} & \underline{57.8} & \underline{60.6} & \underline{58.9} & \underline{99.3} \\
AOT~\cite{li2026token} & CVPR'26 & 20\% & 57.2 & 56.2 & 60.9 & \textbf{58.1} & \textbf{61.3} & 58.7 & 98.9 \\
\rowcolor[RGB]{234,234,255}
\textbf{COMPACT} & -- & 20\% & \textbf{58.4} & \textbf{58.1} & \textbf{63.4} & 57.7 & 60.3 & \textbf{59.6} & \textbf{100.4} \\
\bottomrule
\end{tabular}%
}
\end{table*}
\begin{table*}[t]
\caption{Comparison with state-of-the-art methods on LLaVA-Video-7B. ``Avg. Score'' and ``Avg. \%'' denote the average score across benchmarks and the relative performance to the unpruned model. The best and second-best results are in \textbf{bold} and \underline{underlined}, respectively.}
\label{tab:sup_llava_video}
\centering
\resizebox{\textwidth}{!}{%
\begin{tabular}{l|c|c|ccccc|cc}
\toprule
\multirow{2}{*}{Method}
& \multirow{2}{*}{Source}
& \multirow{2}{*}{\shortstack{Before LLM\\Retained Ratio}}
& \multirow{2}{*}{\shortstack{VideoMME\\$\uparrow$}}
& \multirow{2}{*}{\shortstack{LongVideoBench\\$\uparrow$}}
& \multirow{2}{*}{\shortstack{MLVU\\$\uparrow$}}
& \multirow{2}{*}{\shortstack{MVBench\\$\uparrow$}}
& \multirow{2}{*}{\shortstack{EgoSchema\\$\uparrow$}}
& \multicolumn{2}{|c}{Avg. $\uparrow$} \\
& & & & & & & & Score & \% \\
\midrule
\rowcolor[RGB]{235,235,235}
LLaVA-Video-7B~\cite{zhang2025llava} & -- & 100\% & 64.3 & 58.9 & 68.6 & 60.4 & 57.2 & 61.9 & 100.0 \\
\midrule
VisionZip~\cite{yang2025visionzip} & CVPR'25 & 25\% & 60.7 & 54.7 & 65.9 & 56.7 & 54.7 & 58.5 & 94.6 \\
EarlyTom~\cite{wang2026earlytom} & CVPR'26 & 25\% & \underline{61.9} & \underline{57.3} & 65.0 & \underline{59.1} & \underline{55.3} & 59.7 & 96.5 \\
AOT~\cite{li2026token} & CVPR'26 & 25\% & \textbf{62.4} & 56.2 & \underline{66.2} & 58.8 & \textbf{55.4} & \underline{59.8} & \underline{96.6} \\
\rowcolor[RGB]{234,234,255}
\textbf{COMPACT} & -- & 25\% & 61.7 & \textbf{59.0} & \textbf{67.0} & \textbf{61.2} & 54.0 & \textbf{60.6} & \textbf{97.9} \\
\midrule
VisionZip~\cite{yang2025visionzip} & CVPR'25 & 20\% & \textbf{62.0} & 57.0 & 65.3 & 60.5 & \textbf{55.5} & \underline{60.1} & \underline{97.1} \\
EarlyTom~\cite{wang2026earlytom} & CVPR'26 & 20\% & 61.2 & 54.3 & 63.9 & 59.1 & 54.4 & 58.6 & 94.7 \\
AOT~\cite{li2026token} & CVPR'26 & 20\% & \textbf{62.0} & 56.4 & \underline{65.7} & \underline{61.0} & \underline{55.1} & 60.0 & \underline{97.1} \\
\rowcolor[RGB]{234,234,255}
\textbf{COMPACT} & -- & 20\% & \underline{61.8} & \textbf{58.7} & \textbf{66.7} & \textbf{61.1} & 53.7 & \textbf{60.4} & \textbf{97.6} \\
\bottomrule
\end{tabular}%
}
\end{table*}
\begin{table*}[!ht]
\caption{Comparison with state-of-the-art methods on InternVL3-8B. ``Avg. Score'' and ``Avg. \%'' denote the average score across benchmarks and the relative performance to the unpruned model. The best and second-best results are in \textbf{bold} and \underline{underlined}, respectively.}
\label{tab:sup_internvl3}
\centering
\resizebox{\textwidth}{!}{%
\begin{tabular}{l|c|c|ccccc|cc}
\toprule
\multirow{2}{*}{Method}
& \multirow{2}{*}{Source}
& \multirow{2}{*}{\shortstack{Before LLM\\Retained Ratio}}
& \multirow{2}{*}{\shortstack{VideoMME\\$\uparrow$}}
& \multirow{2}{*}{\shortstack{LongVideoBench\\$\uparrow$}}
& \multirow{2}{*}{\shortstack{MLVU\\$\uparrow$}}
& \multirow{2}{*}{\shortstack{MVBench\\$\uparrow$}}
& \multirow{2}{*}{\shortstack{EgoSchema\\$\uparrow$}}
& \multicolumn{2}{|c}{Avg. $\uparrow$} \\
& & & & & & & & Score & \% \\
\midrule
\rowcolor[RGB]{235,235,235}
InternVL3-8B~\cite{zhu2025internvl3} & -- & 100\% & 64.9 & 58.7 & 68.8 & 74.1 & 62.6 & 65.8 & 100.0 \\
\midrule
VisionZip~\cite{yang2025visionzip} & CVPR'25 & 25\% & \textbf{64.9} & \textbf{58.8} & 68.2 & \textbf{74.3} & \underline{61.2} & \textbf{65.5} & \textbf{99.5} \\
EarlyTom~\cite{wang2026earlytom} & CVPR'26 & 25\% & \underline{64.3} & 58.5 & \underline{68.4} & 73.9 & \textbf{61.3} & \underline{65.3} & 99.2 \\
AOT~\cite{li2026token} & CVPR'26 & 25\% & 61.8 & 55.0 & 65.2 & 73.7 & 59.3 & 63.0 & 95.7 \\
\rowcolor[RGB]{234,234,255}
\textbf{COMPACT} & -- & 25\% & \underline{64.3} & \underline{58.6} & \textbf{68.5} & \underline{74.2} & 61.0 & \underline{65.3} & \underline{99.3} \\
\midrule
VisionZip~\cite{yang2025visionzip} & CVPR'25 & 20\% & \textbf{64.1} & 56.7 & 67.8 & \textbf{74.0} & \textbf{61.0} & \underline{64.7} & \underline{98.4} \\
EarlyTom~\cite{wang2026earlytom} & CVPR'26 & 20\% & 63.5 & \textbf{57.7} & \underline{68.1} & \underline{73.8} & \textbf{61.0} & \textbf{64.8} & \textbf{98.5} \\
AOT~\cite{li2026token} & CVPR'26 & 20\% & 61.4 & 55.1 & 64.2 & 73.0 & 58.7 & 62.5 & 94.9 \\
\rowcolor[RGB]{234,234,255}
\textbf{COMPACT} & -- & 20\% & \underline{64.0} & \underline{57.4} & \textbf{68.3} & \underline{73.8} & \underline{60.2} & \textbf{64.8} & \underline{98.4} \\
\bottomrule
\end{tabular}%
}
\end{table*}

\section{Additional Experimental Results}
\label{sec:more_experimental_results}

We provide additional experiments to further evaluate the performance and computational efficiency of COMPACT across different Video LLM architectures and visual token retention ratios.

\subsection{Additional Performance Comparisons}
We provide additional performance comparisons between COMPACT and recent visual token compression methods at 25\% and 20\% visual token retention ratios on LLaVA-OneVision-7B~\cite{li2025llava}, LLaVA-Video-7B~\cite{zhang2025llava}, and InternVL3-8B~\cite{zhu2025internvl3} across five video understanding benchmarks. Tables~\ref{tab:sup_llava_ov}, \ref{tab:sup_llava_video}, and \ref{tab:sup_internvl3} summarize the results on the three backbones, respectively, including individual benchmark scores, average scores, and relative performance compared with the corresponding uncompressed models.

As shown in the tables, COMPACT achieves the highest average scores at both retention ratios on LLaVA-OneVision-7B and LLaVA-Video-7B. Specifically, on LLaVA-OneVision-7B, COMPACT obtains average scores of 59.5 and 59.6 at 25\% and 20\% retention, respectively, slightly surpassing the uncompressed model's average score of 59.3. On LLaVA-Video-7B, it achieves average scores of 60.6 and 60.4 at the corresponding retention ratios, consistently outperforming the compared compression methods. On InternVL3-8B, COMPACT also maintains competitive performance, achieving average scores of 65.3 and 64.8 at 25\% and 20\% retention, respectively. These results demonstrate the effectiveness of COMPACT across different Video LLM architectures and compression levels.

\begin{table*}[t]
\centering
\begin{minipage}[t]{0.49\textwidth}
\centering
\caption{Runtime breakdown on LLaVA-OneVision-7B.}
\label{tab:sup_runtime_llava_ov}
\resizebox{\linewidth}{!}{%
\begin{tabular}{l|c|ccc}
\toprule
Method
& \shortstack{Retained\\Ratio}
& \shortstack{Clustering Core\\(ms) $\downarrow$}
& \shortstack{Compression\\(ms) $\downarrow$}
& \shortstack{LLM Prefill\\(ms) $\downarrow$} \\
\midrule
\rowcolor[RGB]{235,235,235}
LLaVA-OV-7B~\cite{li2025llava}
& 100\% & -- & -- & 559.0 \\
\midrule
AOT~\cite{li2026token}
& 25\% & -- & 301.0 & 144.2 \\
\rowcolor[RGB]{234,234,255}
\textbf{COMPACT}
& 25\% & 158.6 & 167.5 & 144.9 \\
\midrule
AOT~\cite{li2026token}
& 20\% & -- & 273.1 & 116.8 \\
\rowcolor[RGB]{234,234,255}
\textbf{COMPACT}
& 20\% & 133.8 & 142.6 & 121.7 \\
\midrule
AOT~\cite{li2026token}
& 15\% & -- & 251.8 & 93.2 \\
\rowcolor[RGB]{234,234,255}
\textbf{COMPACT}
& 15\% & 99.1 & 108.9 & 93.1 \\
\midrule
AOT~\cite{li2026token}
& 10\% & -- & 211.7 & 70.7 \\
\rowcolor[RGB]{234,234,255}
\textbf{COMPACT}
& 10\% & 67.7 & 74.6 & 70.6 \\
\midrule
AOT~\cite{li2026token}
& 5\% & -- & 173.0 & 47.4 \\
\rowcolor[RGB]{234,234,255}
\textbf{COMPACT}
& 5\% & 37.6 & 40.8 & 47.9 \\
\bottomrule
\end{tabular}%
}
\end{minipage}
\hfill
\begin{minipage}[t]{0.49\textwidth}
\centering
\caption{Runtime breakdown on LLaVA-Video-7B.}
\label{tab:sup_runtime_llava_video}
\resizebox{\linewidth}{!}{%
\begin{tabular}{l|c|ccc}
\toprule
Method
& \shortstack{Retained\\Ratio}
& \shortstack{Clustering Core\\(ms) $\downarrow$}
& \shortstack{Compression\\(ms) $\downarrow$}
& \shortstack{LLM Prefill\\(ms) $\downarrow$} \\
\midrule
\rowcolor[RGB]{235,235,235}
LLaVA-Video-7B~\cite{zhang2025llava}
& 100\% & -- & -- & 1076.4 \\
\midrule
AOT~\cite{li2026token}
& 25\% & -- & 448.7 & 311.4 \\
\rowcolor[RGB]{234,234,255}
\textbf{COMPACT}
& 25\% & 375.6 & 466.6 & 311.3 \\
\midrule
AOT~\cite{li2026token}
& 20\% & -- & 410.8 & 265.9 \\
\rowcolor[RGB]{234,234,255}
\textbf{COMPACT}
& 20\% & 320.1 & 406.8 & 265.4 \\
\midrule
AOT~\cite{li2026token}
& 15\% & -- & 350.2 & 219.2 \\
\rowcolor[RGB]{234,234,255}
\textbf{COMPACT}
& 15\% & 242.3 & 325.7 & 218.6 \\
\midrule
AOT~\cite{li2026token}
& 10\% & -- & 285.7 & 164.3 \\
\rowcolor[RGB]{234,234,255}
\textbf{COMPACT}
& 10\% & 164.3 & 236.0 & 175.8 \\
\midrule
AOT~\cite{li2026token}
& 5\% & -- & 217.4 & 123.7 \\
\rowcolor[RGB]{234,234,255}
\textbf{COMPACT}
& 5\% & 83.7 & 146.7 & 123.5 \\
\bottomrule
\end{tabular}%
}
\end{minipage}
\end{table*}

\subsection{Computational Efficiency Analysis}
We further evaluate the computational efficiency of COMPACT by comparing its runtime breakdown with AOT on LLaVA-OneVision-7B and LLaVA-Video-7B. Tables~\ref{tab:sup_runtime_llava_ov} and \ref{tab:sup_runtime_llava_video} report three metrics: clustering core time, total compression time, and LLM prefill time. All measurements are conducted on a single NVIDIA A800 80GB GPU with a batch size of 1 and BF16 precision, using the same video--prompt pair. We adopt CUDA-synchronized wall-clock timing and report the average over ten measured runs following two warm-up runs. Specifically, clustering core time covers feature normalization, prototype initialization, and iterative token assignment and centroid updates. Total compression time includes the clustering core, positional encoding, learnable prototype refinement, soft grouping, and visual sequence preparation, excluding visual encoding, multimodal projection, and spatial pooling. The clustering core is included in the total compression time but reported separately for detailed analysis. LLM prefill time measures the initial language-model forward pass, including the language-model head, with generation limited to one token and no subsequent autoregressive decoding. Model loading, video decoding, preprocessing, and tokenization are excluded from all measurements.

On LLaVA-OneVision-7B, COMPACT consistently achieves lower total compression time than AOT across all evaluated retention ratios, reducing the overhead from 173.0\,ms to 40.8\,ms at 5\% retention and from 301.0\,ms to 167.5\,ms at 25\% retention. On LLaVA-Video-7B, COMPACT also achieves lower compression time at retention ratios of 5\%--20\%. Regarding LLM prefill, COMPACT substantially reduces processing time compared with the uncompressed models. At 5\% retention, prefill time decreases from 559.0\,ms to 47.9\,ms on LLaVA-OneVision-7B and from 1076.4\,ms to 123.5\,ms on LLaVA-Video-7B. These results demonstrate that COMPACT achieves favorable compression efficiency while substantially reducing the computational overhead of processing long visual sequences.

\section{Additional Implementation Details}
\label{sec:implementation-details}
We provide additional implementation details, including the reproduction settings of compared methods, the architecture and implementation of COMPACT, and the training configurations used in our experiments.

\subsection{Reproduction Details of Compared Methods}
\label{sec:baseline-implementation}
We consider four representative visual token compression methods as baselines: VisionZip~\cite{yang2025visionzip}, FastVID~\cite{shen2025fastvid}, EarlyTom~\cite{wang2026earlytom}, and AOT~\cite{li2026token}. For LLaVA-OneVision-7B and LLaVA-Video-7B, we use the results reported in the original papers whenever available. For settings not covered by the original papers, including additional token retention ratios and benchmark configurations, we obtain results using the corresponding officially released implementations. We follow the original compression procedures and hyperparameter configurations of each method, adjusting only the budget-related parameters when evaluating additional retention ratios.

For InternVL3-8B, we adapt the released implementations of these methods to the corresponding backbone. The adaptations account for model-specific interfaces, feature dimensions, visual-token layouts, and token insertion into the LLM. We preserve the original token selection, matching, merging, and aggregation operations, together with their corresponding hyperparameter settings.

To ensure fair comparisons, we maintain consistent model checkpoints, frame sampling strategies, text prompts, generation configurations, and benchmark evaluation protocols across methods using the same backbone, while retaining the method-specific processing procedures required by their original implementations. Unless otherwise specified, token retention is measured based on the number of visual content tokens immediately before LLM input. Model-specific structural tokens, including video newline and spatial-grid newline tokens, are excluded from the visual token budget.

We follow the native visual input processing procedures of each backbone, including frame sampling, visual encoding, projection, and spatial pooling. These settings are kept consistent across different compression methods evaluated on the same backbone, while the compression operations are applied at their respective stages.

For LLaVA-OneVision-7B, we uniformly sample 32 frames. Following its native stride-2 bilinear pooling, each frame produces a $14\times14$ visual-token grid, resulting in
\begin{equation}
N = 32\times14\times14 = 6{,}272
\end{equation}
visual content tokens per video.

For LLaVA-Video-7B, we uniformly sample 64 frames following the fixed-64-frame evaluation protocol. After native stride-2 average pooling, each frame produces a $13\times13$ visual-token grid, yielding
\begin{equation}
N = 64\times13\times13 = 10{,}816
\end{equation}
visual content tokens. The 832 spatial-grid newline tokens are preserved and counted separately from the visual content-token budget.

For InternVL3-8B, we uniformly sample 32 frames and follow its native dynamic image-tiling procedure, using 448-pixel tile inputs with \texttt{max\_num=12} and \texttt{total\_max\_num=64}. Native thumbnail tiles are retained. Each tile produces a $16\times16$ projected visual-token grid. Since the number of tiles varies across videos, the total number of visual content tokens is sample-dependent, and the target compression budget is determined by the actual token count of each video.

For all three backbones, given $N$ visual content tokens and a target retention ratio $r$, the nominal output token budget is determined by
\begin{equation}
B = \max\left(1,\operatorname{round}(rN)\right),
\end{equation}
where $B$ denotes the target number of compressed visual tokens. Structural tokens are excluded from this budget. The actual output token count follows each method's native budget constraints.

The implementation details of individual baselines are provided below.

\begin{itemize}

\item \textbf{VisionZip (CVPR 2025).}
Following VisionZip~\cite{yang2025visionzip}, we extract token importance scores and key features from the penultimate vision encoder layer. Key features are averaged across attention heads following the released implementation. Attention-based selection is used to identify dominant tokens, while the remaining tokens are assigned to selected representatives to construct contextual tokens through feature merging. We preserve the original allocation between dominant and contextual tokens. For multi-frame or multi-tile inputs, the video-level token budget is distributed using integer largest-remainder allocation, ensuring non-empty dominant and contextual branches whenever the budget permits.

\item \textbf{FastVID (NeurIPS 2025).}
We reproduce FastVID~\cite{shen2025fastvid} using its released dynamic segmentation, spatiotemporal pruning, and dynamic token merging components. Specifically, we set $c=8$ and $\tau=0.9$ for dynamic segmentation, $d=0.4$ for spatiotemporal pruning, and $p=4$ and $\beta=0.6$ for dynamic token merging. We retain the original segment-wise allocation and token construction and merging procedures for salient and contextual tokens, including their integer-rounding operations. FastVID is excluded from the 1\% retention setting because its structural constraints require more tokens than the available budget.

\item \textbf{EarlyTom (CVPR 2026).}
We use the officially released EarlyTom~\cite{wang2026earlytom} implementation with its task-specific configurations. The exponential moving average coefficient is set to $0.9$, and the temporal matching range is set to $M=6$. The pruning threshold and vision-layer schedule follow the corresponding benchmark and retention-ratio configurations. For additional low-retention settings, we adopt the released low-budget configuration and adjust the target token budget accordingly. We retain the original internal token reduction at LLM layer 18 with a retention ratio of $0.5$. For vision encoders with different depths, the pruning locations are mapped to their corresponding relative depths while preserving the original operation order and pruning rules.

\item \textbf{AOT (CVPR 2026).}
We reproduce AOT~\cite{li2026token} using its original intra-frame and inter-frame optimal transport operations. For inter-frame transport, the Sinkhorn regularization coefficient is set to $0.1$, with a maximum of 100 iterations, and both transport-cost scaling factors are set to $1.0$. We preserve the backbone-specific keep ratios, local and global anchor construction, window rounding, and temporal aggregation procedures. AOT is excluded from the 1\% retention setting because its structural constraints require more tokens than the available budget.

\end{itemize}

\subsection{Implementation Details of COMPACT}
\label{sec:compact-implementation}

COMPACT operates on projected visual tokens after the native visual encoding, projection, and spatial pooling operations. For InternVL3-8B, tiles from the same frame share a temporal coordinate, while their spatial coordinates are mapped to the corresponding locations in the original frame. The output tokens are ordered according to the earliest original token in each initial cluster, and this ordering remains fixed during subsequent learnable processing.

\noindent \textbf{Network Architecture.}
The projected visual features have a dimension of $D=3584$ across all three backbones. We use a shared internal feature dimension of $d=256$ and eight attention heads, resulting in 32 dimensions per head. The prototype refinement module consists of one self-attention block, one cross-attention block, and a feed-forward network with a hidden dimension of 512 and GELU activation. All blocks adopt pre-layer normalization and residual connections, with dropout set to zero. The same compression-module architecture is used for all three backbones, which are trained independently.

\noindent \textbf{Numerical and Runtime Implementation.}
We set the 3D RoPE frequency base to $\theta=10{,}000$. Initial clustering is performed using cosine similarity with deterministic farthest-point initialization. The assignment and center-update procedures run for at most ten iterations and terminate early upon convergence. FP32 computation is used for assignment and aggregation normalization to improve numerical stability.
Final feature aggregation operates directly on the original, unrotated visual features. To reduce intermediate memory consumption during training, aggregation is computed head by head with activation checkpointing. The same continuous learned grouping and aggregation operations are used during training and inference.

\begin{table}[t]
\centering
\caption{Training hyperparameters of COMPACT. The same optimization configuration is used across all three backbones. Sampled frame counts are listed in the order of LLaVA-OneVision-7B, LLaVA-Video-7B, and InternVL3-8B.}
\label{tab:training-hyperparameters}
\resizebox{\columnwidth}{!}{%
\begin{tabular}{lcc}
\toprule
\multirow{2}{*}{Hyperparameter}
& Stage 1 & Stage 2 \\
& Caption & QA \\
\midrule
Learning rate & $2\times10^{-5}$ & $2\times10^{-5}$ \\
Training records & 99,163 & 111,309 \\
Module initialization & Random & Inherit Stage 1 \\
Optimizer state & New & New \\
\midrule
Trainable parameters & \multicolumn{2}{c}{Compression module only} \\
Optimizer & \multicolumn{2}{c}{AdamW} \\
Learning-rate schedule & \multicolumn{2}{c}{Cosine} \\
Warm-up steps & \multicolumn{2}{c}{12} \\
Weight decay & \multicolumn{2}{c}{$0.01$} \\
Gradient clipping & \multicolumn{2}{c}{$1.0$} \\
Epochs per stage & \multicolumn{2}{c}{1} \\
Per-device batch size & \multicolumn{2}{c}{1} \\
Gradient accumulation & \multicolumn{2}{c}{16} \\
Global batch size & \multicolumn{2}{c}{$128$} \\
Precision & \multicolumn{2}{c}{BF16} \\
DeepSpeed & \multicolumn{2}{c}{ZeRO-3} \\
Gradient checkpointing & \multicolumn{2}{c}{Enabled} \\
Frame sampling & \multicolumn{2}{c}{Uniform} \\
Sampled frames & \multicolumn{2}{c}{32 / 64 / 32} \\
Maximum sequence length & \multicolumn{2}{c}{32,768} \\
Random seed & \multicolumn{2}{c}{1234} \\
\bottomrule
\end{tabular}%
}
\end{table}

\noindent \textbf{Training Hyperparameters.}
We adopt the same two-stage training configuration for all three backbones, as summarized in Table~\ref{tab:training-hyperparameters}. Throughout training, the vision encoder, multimodal projector, and LLM remain frozen, while only the attention layer and linear mapping layer are optimized. Each stage is trained for one epoch with a learning rate of $2\times10^{-5}$. Training is conducted on 8 NVIDIA A800 GPUs, with a per-device batch size of one and 16 gradient accumulation steps, resulting in a global batch size of 128. The learnable parameters are randomly initialized in the first stage. In the second stage, they are initialized with the parameters learned in the first stage, while the optimizer and learning-rate scheduler are reinitialized.
Within each training stage, the prepared training records are randomly divided into six groups corresponding to token retention ratios of 1\%, 5\%, 10\%, 15\%, 20\%, and 25\%. The assignment is performed using random seed 1234, with group sizes balanced such that the difference in record counts between any two groups is at most one. Each record is assigned a single retention ratio within a stage. The same prepared training records and ratio assignments are shared across all three backbones.

\section{A Unified View of Visual Token Compression}
\label{sec:unified_assignment}

We establish a unified view of representative visual token compression methods from the perspective of token aggregation. Existing approaches employ different strategies to transform dense visual tokens into compact representations, including token selection, similarity-based merging, and anchor-based aggregation. Despite their different designs, these methods can all be characterized by the contributions of input tokens to compressed outputs, providing a common formulation for understanding their connections to COMPACT.

\noindent\textbf{General Formulation.}
Given dense visual tokens $X=\{x_i\}_{i=1}^{N}$ and a target of $K$ compressed tokens, we formulate token compression as a weighted aggregation operation with a contribution matrix $\Omega\in\mathbb{R}_{\geq0}^{N\times K}$, where $\Omega_{ik}$ represents the contribution of the $i$-th input token to the $k$-th output token. The compressed tokens are computed as:
\begin{equation}
y_k=
\frac{\sum_{i=1}^{N}\Omega_{ik}x_i}
{\sum_{i=1}^{N}\Omega_{ik}},
\qquad
\sum_{i=1}^{N}\Omega_{ik}>0.
\label{eq:unified_weighted_aggregation}
\end{equation}

Under this formulation, different compression methods primarily differ in how their contribution weights are determined. Token selection produces sparse binary weights, similarity-based merging assigns multiple input tokens to shared outputs, and anchor-based aggregation introduces structured weights according to token importance, similarity, or optimal transport.

\noindent\textbf{Token Selection and Pruning.}
DynamicViT~\cite{rao2021dynamicvit} progressively removes redundant tokens using importance scores predicted from intermediate visual features. FastV~\cite{chen2024image} ranks visual tokens according to their average received attention scores at a selected LLM layer and prunes low-scoring tokens in subsequent layers. DivPrune~\cite{alvar2025divprune} formulates token selection as a Max-Min Diversity Problem (MMDP), explicitly maximizing the minimum pairwise distance among retained tokens. Its selection objective is:
\begin{equation}
\mathcal{S}^{*}
=
\arg\max_{\substack{
\mathcal{S}\subseteq\{1,\ldots,N\}\\
|\mathcal{S}|=K
}}
\min_{\substack{
i,j\in\mathcal{S}\\i\neq j
}}
\left(
1-
\frac{x_i^\top x_j}
{\|x_i\|_2\|x_j\|_2}
\right).
\label{eq:divprune_objective}
\end{equation}

Despite their different selection criteria, these methods share the same binary aggregation structure. Let $\mathcal{S}^{(m)}=\{j_1^{(m)},\ldots,j_K^{(m)}\}$ denote the indices selected by method $m$, where $m\in\{\mathrm{DynamicViT},\mathrm{FastV},\mathrm{DivPrune}\}$. Their contribution weights are expressed as:
\begin{equation}
\Omega_{ik}^{(m)}=
\begin{cases}
1, & i=j_k^{(m)},\\
0, & \text{otherwise}.
\end{cases}
\label{eq:selection_weights}
\end{equation}

Thus, DynamicViT represents an importance-guided hard-selection special case of COMPACT, where the contribution matrix is determined by learned token importance scores. FastV represents an attention-guided hard-selection special case, with retained tokens determined by their relevance to text queries. DivPrune represents a diversity-constrained hard-selection special case, where the selected tokens are determined by the max-min diversity objective. All three methods produce singleton outputs through binary token-to-output assignments.

\noindent\textbf{Similarity-Based Token Merging.}
Token Merging~\cite{bolya2023token} progressively merges similar tokens through bipartite matching. Input tokens are divided into source and target sets, and source tokens are matched to their most similar eligible targets using feature similarity. The selected matching pairs form merging groups, while unmatched tokens remain unchanged. To preserve the contributions of previously merged tokens, Token Merging maintains an accumulated token size $\nu_i$. Let $\mathcal{C}_k$ denote the group associated with the $k$-th output token. Its aggregation weights are:
\begin{equation}
\Omega_{ik}^{\mathrm{TokenMerging}}=
\begin{cases}
\nu_i, & i\in\mathcal{C}_k,\\
0, & \text{otherwise}.
\end{cases}
\label{eq:tome_weights}
\end{equation}

Token Merging therefore represents a hard-assignment, size-weighted special case of COMPACT. Each input token belongs to a single output group, while its accumulated token size determines its relative contribution to the merged representation.

VisionZip~\cite{yang2025visionzip} combines attention-guided dominant token selection with similarity-based contextual token merging. It first retains dominant tokens according to their visual attention scores and then divides the remaining tokens into target and merge sets. Each merge token is assigned to its most similar target based on key-feature similarity, and tokens within each resulting group are averaged. Let $\mathcal{K}_d$ and $\mathcal{K}_c$ denote the dominant-token and contextual-token output sets, respectively. The corresponding aggregation weights are:
\begin{equation}
\Omega_{ik}^{\mathrm{VisionZip}}=
\begin{cases}
1, & k\in\mathcal{K}_d,\ i=j_k,\\
1, & k\in\mathcal{K}_c,\ i\in\mathcal{C}_k^{\mathrm{VisionZip}},\\
0, & \text{otherwise}.
\end{cases}
\label{eq:visionzip_weights}
\end{equation}
Here, each contextual group $\mathcal{C}_k^{\mathrm{VisionZip}}$ contains its target token and all merge tokens assigned to it.

VisionZip represents a hybrid selection-and-merging special case of COMPACT. Its dominant tokens correspond to singleton outputs, while contextual tokens form hard-assigned groups with uniform aggregation weights. Thus, its two compression operations are jointly characterized by a binary contribution matrix.

FrameFusion~\cite{fu2025framefusion} extends similarity-based merging to video inputs by identifying redundant tokens at corresponding spatial positions across adjacent frames. Specifically, it computes cosine similarity between corresponding tokens in consecutive frames and progressively merges tokens whose similarity exceeds a predefined threshold. The resulting groups are transitively constructed across frames, and their features are averaged and assigned to the earliest token in each group. Let $\mathcal{C}_k^{\mathrm{FrameFusion}}$ denote the resulting merging group. Its aggregation weights are:
\begin{equation}
\Omega_{ik}^{\mathrm{FrameFusion}}=
\begin{cases}
1, & i\in\mathcal{C}_k^{\mathrm{FrameFusion}},\\
0, & \text{otherwise}.
\end{cases}
\label{eq:framefusion_weights}
\end{equation}

FrameFusion represents a temporally constrained hard-grouping special case of COMPACT, where spatially aligned tokens from adjacent frames are assigned to the same output with equal contributions. Its subsequent cumulative-attention-based pruning further corresponds to the binary selection formulation in Eq.~\ref{eq:selection_weights}.

\noindent\textbf{Anchor-Based and Spatiotemporal Aggregation.}
FastVID~\cite{shen2025fastvid} identifies representative anchors through spatiotemporal density estimation and assigns redundant tokens to their nearest anchors based on cosine similarity. Its Density-based Token Merging module performs anchor-centric aggregation within temporally segmented video sequences. Let $a_k$ denote the anchor of the $k$-th merging group $\mathcal{C}_k$, and let $\mathcal{N}_k=\mathcal{C}_k\setminus\{a_k\}$ denote its associated non-anchor tokens. Its aggregation weights are:
\begin{equation}
\Omega_{ik}^{\mathrm{FastVID}}=
\begin{cases}
\beta, & i=a_k,\\
\dfrac{1-\beta}{|\mathcal{N}_k|},
& i\in\mathcal{N}_k,\\
0, & \text{otherwise},
\end{cases}
\label{eq:fastvid_weights}
\end{equation}
where $\beta\in[0,1]$ controls the relative contribution of the anchor.

FastVID represents an anchor-biased hard-grouping special case of COMPACT. Its contribution matrix assigns each input token to a single anchor group and explicitly specifies the relative weights of the anchor and its associated tokens. Its attention-based token selection branch additionally follows the singleton-selection formulation in Eq.~\ref{eq:selection_weights}.

AOT~\cite{li2026token} employs attention-guided anchors and local-global optimal transport to aggregate unselected tokens into representative visual features. It performs token aggregation at both intra-frame and inter-frame levels.

For intra-frame aggregation, let $\mathcal{A}=\{a_j\}_{j=1}^{M}$ denote the selected anchors and $\mathcal{U}=\{u_i\}_{i=1}^{N-M}$ denote the unselected tokens. Given the optimal transport plan $T_{\mathrm{intra}}^{*}$, the aggregation weights are:
\begin{equation}
\Omega_{vj}^{\mathrm{intra}}=
\begin{cases}
1, & v=a_j,\\
\lambda_{\mathrm{intra}}T_{ij}^{*}, & v=u_i,\\
0, & \text{otherwise},
\end{cases}
\label{eq:aot_intra_weights}
\end{equation}
where $\lambda_{\mathrm{intra}}$ controls the contribution of unselected tokens. Accordingly, each anchor is updated as:
\begin{equation}
\widetilde{a}_j=
\frac{
a_j+\lambda_{\mathrm{intra}}
\sum_{i=1}^{N-M}T_{ij}^{*}u_i
}{
1+\lambda_{\mathrm{intra}}
\sum_{i=1}^{N-M}T_{ij}^{*}
}.
\label{eq:aot_intra_update}
\end{equation}

The intra-frame operation of AOT represents a transport-weighted anchor-aggregation special case of COMPACT. The optimal transport plan determines the contributions of contextual tokens to each anchor, while the original anchor retains its unit contribution.

For inter-frame aggregation, AOT uses the first frame in each temporal clip as the initial anchor set. Given the optimal transport plan $T_{\mathrm{inter}}^{(\ell)*}$ between the current frame tokens and the existing clip anchors, it first computes row-normalized assignment probabilities:
\begin{equation}
p_{ij}^{(\ell)}
=
\frac{T_{ij}^{(\ell)*}}
{\sum_{j'}T_{ij'}^{(\ell)*}},
\qquad
q_i^{(\ell)}=\max_j p_{ij}^{(\ell)}.
\label{eq:aot_inter_assignment}
\end{equation}

Tokens satisfying $q_i^{(\ell)}\geq\tau$ are aggregated into the clip anchors, with the corresponding weights defined as:
\begin{equation}
\Omega_{vj}^{\mathrm{inter}}=
\begin{cases}
1, & v=a_j^{(\ell-1)},\\
\lambda_{\mathrm{inter}}p_{ij}^{(\ell)},
& \substack{v=s_i^{(\ell)},\\ q_i^{(\ell)}\geq\tau},\\
0, & \text{otherwise}.
\end{cases}
\label{eq:aot_inter_weights}
\end{equation}
where $\lambda_{\mathrm{inter}}$ controls the aggregation strength and $\tau$ determines the assignment threshold. The resulting anchor update is:
\begin{equation}
a_j^{(\ell)}
=
\frac{
a_j^{(\ell-1)}
+
\lambda_{\mathrm{inter}}
\sum_{i:q_i^{(\ell)}\geq\tau}
p_{ij}^{(\ell)}s_i^{(\ell)}
}{
1+
\lambda_{\mathrm{inter}}
\sum_{i:q_i^{(\ell)}\geq\tau}
p_{ij}^{(\ell)}
}.
\label{eq:aot_inter_update}
\end{equation}

Tokens with $q_i^{(\ell)}<\tau$ are retained as individual outputs. The inter-frame operation therefore represents a confidence-gated transport-aggregation special case of COMPACT, combining soft token-to-anchor contributions with hard selection of tokens that preserve distinct temporal information.

EarlyTom~\cite{wang2026earlytom} reduces temporal redundancy by merging adjacent frames inside the vision encoder, followed by spatial token selection. Let $s_t$ denote the similarity between frames $t$ and $t+1$, and let $\pi(t,p_k)$ denote the index of the token at spatial position $p_k$ in frame $t$. For eligible middle-frame pairs, merging is performed when $s_t$ exceeds a predefined threshold and is greater than $s_{t+1}$. The weighted aggregation is defined as:
\begin{equation}
\Omega_{ik}^{\mathrm{EarlyTom}}=
\begin{cases}
s_t, & i=\pi(t,p_k),\\
s_{t+1}, & i=\pi(t+1,p_k),\\
0, & \text{otherwise}.
\end{cases}
\label{eq:earlytom_weights}
\end{equation}

Consequently, the merged representation at spatial position $p_k$ is:
\begin{equation}
y_k=
\frac{
s_t x_{\pi(t,p_k)}
+
s_{t+1}x_{\pi(t+1,p_k)}
}{
s_t+s_{t+1}
}.
\label{eq:earlytom_update}
\end{equation}

EarlyTom represents a temporally local, similarity-weighted aggregation special case of COMPACT. Each output combines two spatially aligned tokens from adjacent frames, with their relative contributions determined by frame similarity. Its subsequent spatial token selection follows the binary selection formulation in Eq.~\ref{eq:selection_weights}.

\noindent\textbf{COMPACT: General and Adaptive Aggregation.}
The above methods instantiate different structured forms of the contribution matrix, including binary selection, hard similarity-based grouping, accumulated-size weighting, anchor-biased aggregation, optimal transport, and similarity-weighted temporal fusion. COMPACT provides a unified formulation that accommodates these different aggregation patterns through prototype-guided soft assignment.

Specifically, given input tokens $X$ and $K$ prototypes, let $s_{ik}(\theta)$ denote the compatibility score between the $i$-th input token and the $k$-th prototype. The contribution weights are defined as:
\begin{equation}
\Omega_{ik}^{\mathrm{COMPACT}}=
\frac{\exp(s_{ik}(\theta))}
{\sum_{j=1}^{K}\exp(s_{ij}(\theta))}.
\label{eq:compact_weights}
\end{equation}

The resulting weights are used in Eq.~\ref{eq:unified_weighted_aggregation} to generate compressed visual representations. Through differentiable soft assignments, COMPACT allows each input token to contribute to multiple output prototypes with continuously adjusted weights. Prototype-guided initialization and subsequent refinement adapt token assignments to visual content and downstream task objectives.

Overall, the representative methods can be understood as specific instances of the weighted aggregation operation underlying COMPACT, each imposing a particular structure or constraint on the contribution matrix. COMPACT extends this formulation by learning token-to-prototype compatibility, providing a unified and adaptive approach to visual token compression. Table~\ref{tab:unified_assignment} summarizes the aggregation formulations and corresponding weight determination mechanisms of these methods.

\begin{table*}[t]
\centering
\caption{A unified weighted-aggregation view of representative visual token compression methods. Each contribution matrix $\Omega$ specifies the contribution of input tokens to compressed outputs through Eq.~\ref{eq:unified_weighted_aggregation}. For multi-stage methods, the table presents the corresponding aggregation operations. For DynamicViT, FastV, and DivPrune, $m$ indexes the method and $j_k^{(m)}$ denotes a selected input index. For Token Merging, $\mathcal{C}_k$ denotes a merging group and $\nu_i$ the accumulated token size. For VisionZip, $\mathcal{K}_d$ and $\mathcal{K}_c$ index dominant and contextual outputs, respectively, with each contextual group containing its target and assigned merge tokens. For FrameFusion, $\mathcal{C}_k^{\mathrm{FrameFusion}}$ contains the original tokens represented by a surviving transitive merging group. For FastVID, $\mathcal{K}_m$ and $\mathcal{K}_s$ index density-merged and attention-selected outputs, respectively; $a_k$ denotes an anchor, $\mathcal{N}_k$ its associated non-anchor tokens, and $\beta$ controls the anchor contribution. For AOT, $a_j$ and $u_i$ denote anchors and unselected tokens, $T_{ij}^{*}$ and $p_{ij}^{(\ell)}$ denote intra- and inter-frame transport assignments, and $q_i^{(\ell)}$ and $\tau$ define the inter-frame confidence threshold. $\mathcal{K}_a$ and $\mathcal{K}_s$ index clip-anchor and singleton outputs, respectively. The coefficients $\lambda_{\mathrm{intra}}$ and $\lambda_{\mathrm{inter}}$ control aggregation strength. For EarlyTom, $\pi(t,p_k)$ denotes the token index at spatial position $p_k$ in frame $t$, and $s_t$ denotes adjacent-frame similarity. For COMPACT, $s_{ik}(\theta)$ denotes the compatibility score between input token $i$ and prototype $k$.}
\label{tab:unified_assignment}
\small
\resizebox{\textwidth}{!}{%
\begin{tabular}{
@{}
>{\raggedright\arraybackslash}m{0.18\textwidth}
>{\centering\arraybackslash}m{0.37\textwidth}
>{\raggedright\arraybackslash}m{0.40\textwidth}
@{}
}
\toprule

\multicolumn{1}{l}{\textbf{Method}} &
\multicolumn{1}{c}{\textbf{Aggregation Weights}} &
\multicolumn{1}{l}{\textbf{Weight Determination}} \\

\midrule

\multicolumn{3}{l}{\textit{Token Selection and Pruning}} \\
\midrule

DynamicViT~\cite{rao2021dynamicvit},
FastV~\cite{chen2024image},
DivPrune~\cite{alvar2025divprune}
&
$\displaystyle
\Omega_{ik}^{(m)}=
\begin{cases}
1, & i=j_k^{(m)},\\
0, & \text{otherwise}
\end{cases}$
&
Learned importance prediction in DynamicViT, text-to-visual attention ranking in FastV, and Max-Min diversity optimization in DivPrune. \\

\midrule

\multicolumn{3}{l}{\textit{Similarity-Based Token Merging}} \\
\midrule

ToMe~\cite{bolya2023token}
&
$\displaystyle
\Omega_{ik}^{\mathrm{ToMe}}=
\begin{cases}
\nu_i, & i\in\mathcal{C}_k,\\
0, & \text{otherwise}
\end{cases}$
&
Bipartite similarity matching with accumulated token-size-weighted aggregation. \\

VisionZip~\cite{yang2025visionzip}
&
$\displaystyle
\Omega_{ik}^{\mathrm{VisionZip}}=
\begin{cases}
1, & k\in\mathcal{K}_d,\ i=j_k,\\
1, & k\in\mathcal{K}_c,\ i\in\mathcal{C}_k^{\mathrm{VisionZip}},\\
0, & \text{otherwise}
\end{cases}$
&
Attention-guided dominant token selection and similarity-based averaging of contextual token groups. \\

FrameFusion~\cite{fu2025framefusion}
&
$\displaystyle
\Omega_{ik}^{\mathrm{FrameFusion}}=
\begin{cases}
1, & i\in\mathcal{C}_k^{\mathrm{FrameFusion}},\\
0, & \text{otherwise}
\end{cases}$
&
Transitive similarity-based merging of spatially aligned tokens across adjacent frames, followed by cumulative-attention pruning. \\

\midrule

\multicolumn{3}{l}{\textit{Anchor-Based and Spatiotemporal Aggregation}} \\
\midrule

FastVID~\cite{shen2025fastvid}
&
$\displaystyle
\Omega_{ik}^{\mathrm{FastVID}}=
\begin{cases}
\beta, & k\in\mathcal{K}_m,\ i=a_k,\\
\dfrac{1-\beta}{|\mathcal{N}_k|},
& k\in\mathcal{K}_m,\ i\in\mathcal{N}_k,\\
1, & k\in\mathcal{K}_s,\ i=j_k,\\
0, & \text{otherwise}
\end{cases}$
&
Density-peak anchor selection and cosine-similarity assignment, combined with attention-based token selection. \\

AOT~\cite{li2026token}\par (intra-frame)
&
$\displaystyle
\Omega_{ij}^{\mathrm{intra}}=
\begin{cases}
1, & i=a_j,\\
\lambda_{\mathrm{intra}}T_{rj}^{*}, & i=u_r,\\
0, & \text{otherwise}
\end{cases}$
&
Attention-guided local/global anchor selection and optimal-transport-based contextual aggregation within each frame. \\

AOT~\cite{li2026token}\par (inter-frame)
&
$\displaystyle
\Omega_{ij}^{\mathrm{inter}}=
\begin{cases}
1, & j\in\mathcal{K}_a,\ i=a_j^{(\ell-1)},\\
\lambda_{\mathrm{inter}}p_{ij}^{(\ell)},
& \substack{j\in\mathcal{K}_a,\\
i\in\mathcal{I}^{(\ell)},\ q_i^{(\ell)}\geq\tau},\\
1, & j\in\mathcal{K}_s,\ i=j,\ q_i^{(\ell)}<\tau,\\
0, & \text{otherwise}
\end{cases}$
&
Confidence-gated inter-frame optimal transport for updating clip anchors while retaining high-change tokens individually. \\

EarlyTom~\cite{wang2026earlytom}
&
$\displaystyle
\Omega_{ik}^{\mathrm{EarlyTom}}=
\begin{cases}
s_t, & i=\pi(t,p_k),\\
s_{t+1}, & i=\pi(t+1,p_k),\\
0, & \text{otherwise}
\end{cases}$
&
Similarity-weighted merging of spatially aligned tokens across adjacent frames, followed by dynamic/static spatial token selection. \\

\midrule

\textbf{COMPACT (Ours)}
&
$\displaystyle
\Omega_{ik}^{\mathrm{COMPACT}}=
\frac{\exp(s_{ik}(\theta))}
{\sum_{r=1}^{K}\exp(s_{ir}(\theta))}$
&
Prototype-guided soft assignment optimized through downstream task supervision. \\

\bottomrule
\end{tabular}%
}
\end{table*}

\section{Additional Visualizations}
\label{sec:more_visualizations}

We present additional visualizations of the clustering results to illustrate the cross-frame grouping behavior of COMPACT and its clustering patterns under different visual token retention ratios.

\subsection{Cross-Frame Clustering}
\begin{figure*}[t]
\centering
\includegraphics[width=\linewidth]{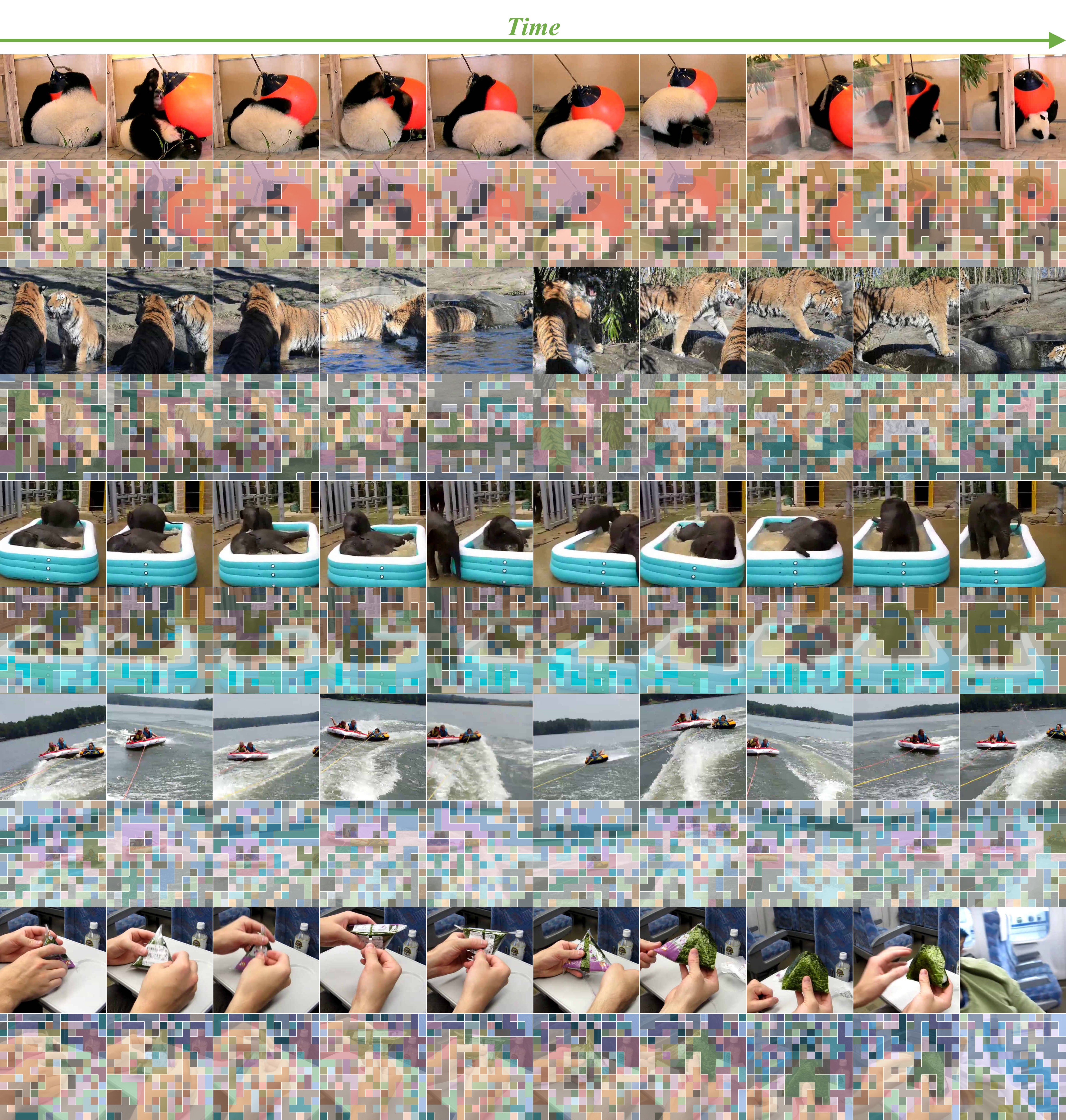}
\caption{Cross-frame clustering visualizations generated by COMPACT at 5\% visual token retention. For each video, the upper row shows sampled frames in temporal order, and the lower row shows the corresponding cluster assignments. Tokens assigned to the same cluster share the same color.}
\label{fig:sup_fig1}
\end{figure*}

Figure~\ref{fig:sup_fig1} presents cross-frame clustering visualizations for five videos with diverse visual contents and temporal dynamics. For each video, the upper row shows sampled frames in temporal order, while the lower row illustrates the corresponding cluster assignments, with tokens belonging to the same cluster represented by the same color. COMPACT effectively groups visually similar regions across different frames while maintaining distinct cluster assignments for different visual patterns. The resulting clusters exhibit coherent structures across temporal sequences, demonstrating the ability of COMPACT to capture cross-frame feature similarities and exploit temporal redundancy.

\subsection{Clustering at Different Retention Ratios}
\begin{figure*}[t]
\centering
\includegraphics[width=\linewidth]{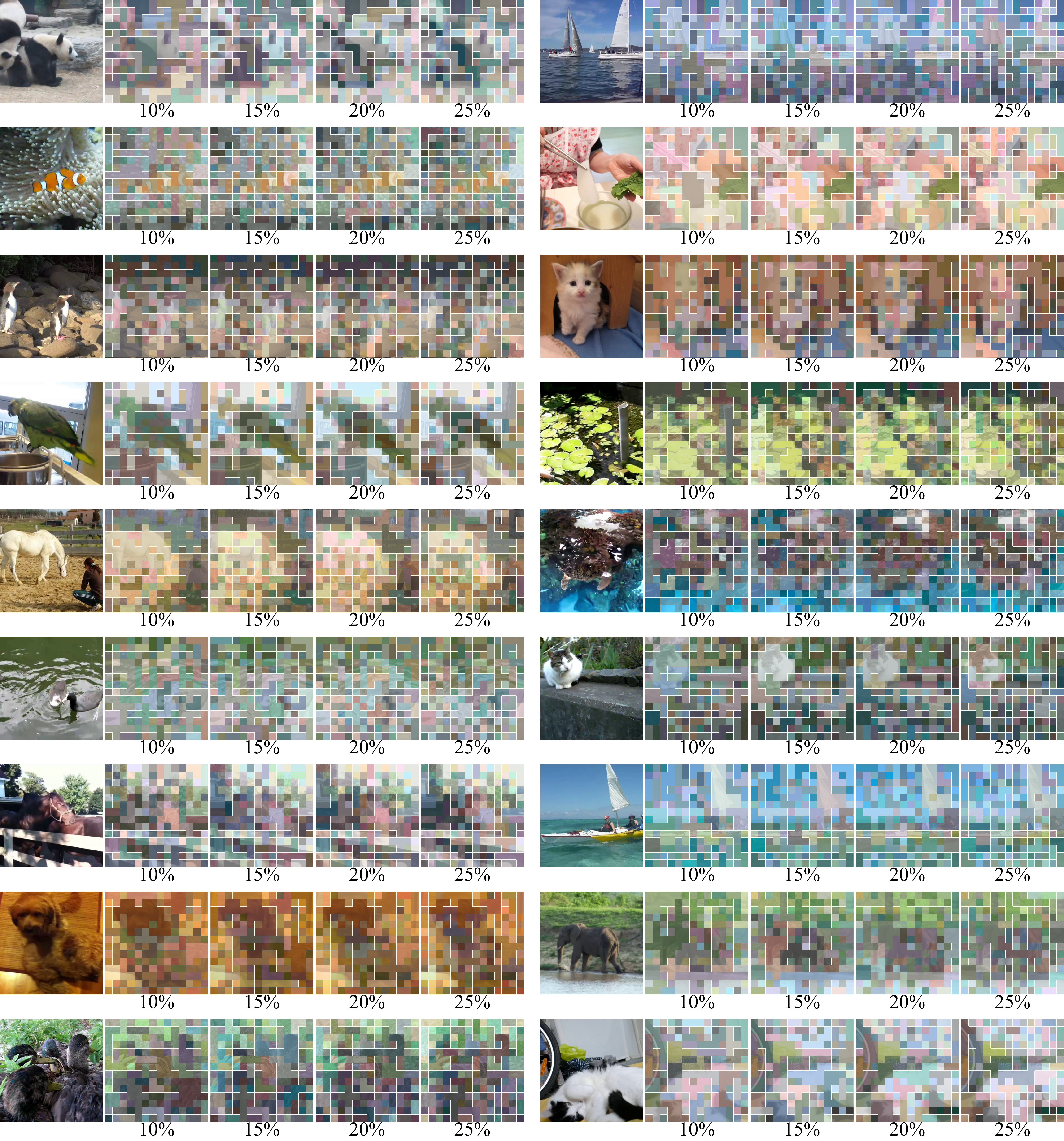}
\caption{Clustering visualizations of individual video frames generated by COMPACT at different visual token retention ratios (10\%, 15\%, 20\%, and 25\%). Each example shows a representative video frame alongside its clustering results at different retention ratios. Tokens assigned to the same cluster share the same color.}
\label{fig:sup_fig2}
\end{figure*}

Figure~\ref{fig:sup_fig2} shows clustering visualizations of individual video frames at different visual token retention ratios (10\%, 15\%, 20\%, and 25\%). As the retention ratio increases, more clusters become available, resulting in finer-grained partitions of the original visual tokens. At lower retention ratios, COMPACT groups broader visual regions into fewer clusters, while higher retention ratios allow more detailed spatial grouping. Across different compression levels and video contents, COMPACT consistently produces coherent clustering structures while preserving the major visual regions.

\end{document}